\documentclass{article}

\usepackage{times}
\usepackage{graphicx} 
\usepackage{subfigure} 

\usepackage{natbib}

\usepackage{algorithm}
\usepackage{algorithmic}

\usepackage{hyperref}
\usepackage{booktabs}
\usepackage{graphicx}
\usepackage{amsmath,amssymb}
\usepackage{array}
\usepackage{multirow}
\usepackage{color,colortbl,soul}
\usepackage[compact]{titlesec}
\usepackage{enumitem}
\usepackage{appendix}
\usepackage{multicol}
\usepackage{graphicx} 
\usepackage{subfigure} 
\usepackage{natbib}
\usepackage{algorithm}
\usepackage{algorithmic}
\usepackage{hyperref}
\usepackage{amsmath}
\usepackage{amsfonts}
\usepackage[all]{xy}

\usepackage{bibunits}

\newcommand{\figref}[1]{Fig.~\ref{#1}}
\newcommand{\tblref}[1]{Table~\ref{#1}}
\newcommand{\sref}[1]{Sect.~\ref{#1}}
\def\eg{\emph{e.g.}}
\def\ie{\emph{i.e.}}
\def\Eg{\emph{E.g.}}

\definecolor{Gray}{gray}{0.95}

\newcommand{\dd}[2]{\frac{\partial{#1}}{\partial{#2}}}

\newcommand{\blue}[1]{{\color{blue}{#1}}}

\newcommand{\expect}[1]{\mathbb{E} \left[ #1 \right]}
\newcommand{\expectx}[2]{\mathbb{E}_{#1} \left[ #2 \right]}

\usepackage[accepted]{icml2017}

\icmltitlerunning{Decoupled Neural Interfaces using Synthetic Gradients}

\begin{document} 

\twocolumn[
\icmltitle{Decoupled Neural Interfaces using Synthetic Gradients}



\icmlsetsymbol{equal}{*}

\begin{icmlauthorlist}
\icmlauthor{Max Jaderberg}{dm}
\icmlauthor{Wojciech Marian Czarnecki}{dm}
\icmlauthor{Simon Osindero}{dm}
\icmlauthor{Oriol Vinyals}{dm}
\icmlauthor{Alex Graves}{dm}
\icmlauthor{David Silver}{dm}
\icmlauthor{Koray Kavukcuoglu}{dm}
\end{icmlauthorlist}

\icmlaffiliation{dm}{DeepMind, London, UK}

\icmlcorrespondingauthor{Max Jaderberg}{jaderberg@google.com}

\icmlkeywords{deep learning, machine learning, ICML}

\vskip 0.3in
]



\printAffiliationsAndNotice{}  


\begin{abstract}
Training directed neural networks typically requires forward-propagating data through a computation graph, followed by backpropagating error signal, to produce weight updates. All layers, or more generally, modules, of the network are therefore locked, in the sense that they must wait for the remainder of the network to execute forwards and propagate error backwards before they can be updated. In this work we break this constraint by decoupling modules by introducing a model of the future computation of the network graph. These models predict what the result of the modelled subgraph will produce using only local information. In particular we focus on modelling error gradients: by using the modelled \emph{synthetic gradient} in place of true backpropagated error gradients we decouple subgraphs, and can update them independently and asynchronously \ie~we realise \emph{decoupled neural interfaces}. We show results for feed-forward models, where every layer is trained asynchronously, recurrent neural networks (RNNs) where predicting one's future gradient extends the time over which the RNN can effectively model, and also a hierarchical RNN system with ticking at different timescales. Finally, we demonstrate that in addition to predicting gradients, the same framework can be used to predict inputs, resulting in models which are decoupled in both the forward and backwards pass -- amounting to independent networks which co-learn such that they can be composed into a single functioning corporation.
\end{abstract}

\section{Introduction}
Each layer (or module) in a directed neural network can be considered a computation step, that transforms its incoming data. These modules are connected via directed edges, creating a forward processing graph which defines the flow of data from the network inputs, through each module, producing network outputs. Defining a loss on outputs allows errors to be generated, and propagated back through the network graph to provide a signal to update each module.

This process results in several forms of \emph{locking}, namely: (i) \emph{Forward Locking} -- no module can process its incoming data before the previous nodes in the directed forward graph have executed; (ii) \emph{Update Locking} -- no module can be updated before all dependent modules have executed in forwards mode; also, in many credit-assignment algorithms (including backpropagation~\citep{Rumelhart86}) we have (iii)~\emph{Backwards Locking} --  no module can be updated before all dependent modules have executed in both forwards mode and backwards mode.


\begin{figure}[t!]
 \centering
 \begin{tabular}{cccc}
 \includegraphics[width=0.4\textwidth]{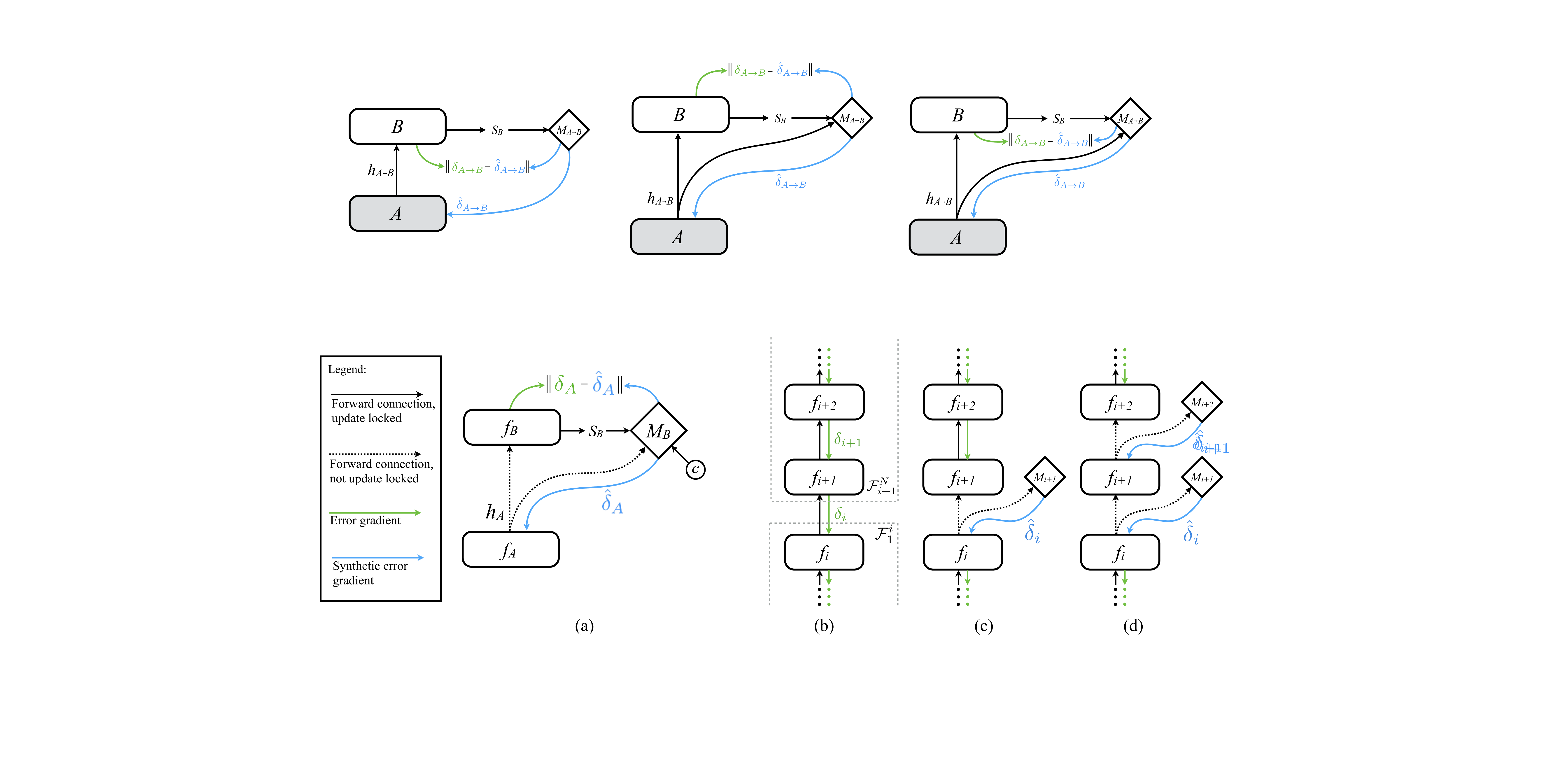}
 \end{tabular}
 \caption{\small General communication protocol between $A$ and $B$. After receiving the message $h_{A}$ from $A$, $B$ can use its model of $A$, $M_{B}$, to send back \emph{synthetic gradients} $\hat{\delta}_{A}$ which are trained to approximate real error gradients $\delta_{A}$. Note that $A$ does not need to wait for any extra computation after itself to get the correct error gradients, hence decoupling the backward computation. The feedback model $M_{B}$ can also be conditioned on any privileged information or context, $c$, available during training such as a label.}
 \label{fig:comms}
 \end{figure}

Forwards, update, and backwards locking constrain us to running and updating neural networks in a sequential, synchronous manner. Though seemingly benign when training simple feed-forward nets, this poses problems when thinking about creating systems of networks acting in multiple environments at different and possibly irregular or asynchronous timescales. For example, in complex systems comprised of multiple asynchronous cooperative modules (or agents), it is undesirable and potentially unfeasible that all networks are update locked. Another example is a distributed model, where part of the model is shared and used by many downstream clients --  all clients must be fully executed and pass error gradients back to the shared model before the model can update, meaning the system trains as fast as the slowest client. The possibility to parallelise training of currently sequential systems could hugely speed up computation time.

The goal of this work is to remove update locking for neural networks. This is achieved by removing backpropagation. To update weights $\theta_i$ of module $i$ we drastically approximate the function implied by backpropagation:
\begin{equation*}
\begin{split}
\frac{\partial L}{\partial \theta_i} & = f_\text{Bprop}((h_i, x_i, y_i, \theta_i),\ldots)\frac{\partial h_i}{\partial \theta_i}
\\ &\simeq~ \hat{f}_\text{Bprop}(h_i)\frac{\partial h_i}{\partial \theta_i}
\end{split}
\end{equation*}
where $h$ are activations, $x$ are inputs, $y$ is supervision, and $L$ is the overall loss to minimise. This leaves dependency only on $h_i$ -- the information local to module $i$.

The premise of this method is based on a simple protocol for learnt communication, allowing neural network modules to interact and be trained without update locking. While the communication protocol is general with respect to the means of generating  a training signal, here we focus on a specific implementation for networks trained with gradient descent -- we replace a standard \emph{neural interface} (a connection between two modules in a neural network) with a Decoupled Neural Interface (DNI). Most simply, when a module (\eg~a layer) sends a message (activations) to another module, there is an associated model which produces a predicted error gradient with respect to the message immediately. The predicted gradient is a function of the message alone; there is no dependence on downstream events, states or losses. The sender can then immediately use these \emph{synthetic gradients} to get an update, without incurring any delay. And by removing update- and backwards locking in this way, we can train networks without a synchronous backward pass. We also show preliminary results that extend this idea to also remove forward locking -- resulting in networks whose modules can also be trained without a synchronous forward pass. When applied to RNNs we show that using synthetic gradients allows RNNs to model much greater time horizons than the limit imposed by truncating backpropagation through time (BPTT). We also show that using synthetic gradients to decouple a system of two RNNs running at different timescales can greatly increase training speed of the faster RNN.

Our synthetic gradient model is most analogous to a value function which is used for gradient ascent~\citep{Baxter00} or critics for training neural networks~\citep{schmidhuber1990networks}. Most other works that aim to remove backpropagation do so with the goal of performing biologically plausible credit assignment, but this doesn't eliminate update locking between layers. \emph{E.g.} target propagation~\citep{Lee15,Bengio14} removes the reliance on passing gradients between layers, by instead generating target activations which should be fitted to. However these targets must still be generated sequentially, propagating backwards through the network and layers are therefore still update- and backwards-locked. Other algorithms remove the backwards locking by allowing loss or rewards to be broadcast directly to each layer -- \eg~REINFORCE~\citep{Williams92} (considering all activations are actions), Kickback~\citep{Balduzzi14}, and Policy Gradient Coagent Networks~\citep{Thomas11}~--~but still remain update locked since they require rewards to be generated by an output (or a global critic). While Real-Time Recurrent Learning~\citep{Williams89} or approximations such as \citep{Ollivier15,Tallec17} may seem a promising way to remove update locking, these methods require maintaining the full (or approximate) gradient of the current state with respect to the parameters. This is inherently not scalable and also requires the optimiser to have global knowledge of the network state. In contrast, by framing the interaction between layers as a local communication problem with DNI, we remove the need for global knowledge of the learning system. Other works such as \citep{Taylor16,Carreira14} allow training of layers in parallel without backpropagation, but in practice are not scalable to more complex and generic network architectures. 



\section{Decoupled Neural Interfaces}\label{sec:communication}

\begin{figure}[t]
\centering
\begin{tabular}{cccc}
\includegraphics[width=0.5\textwidth]{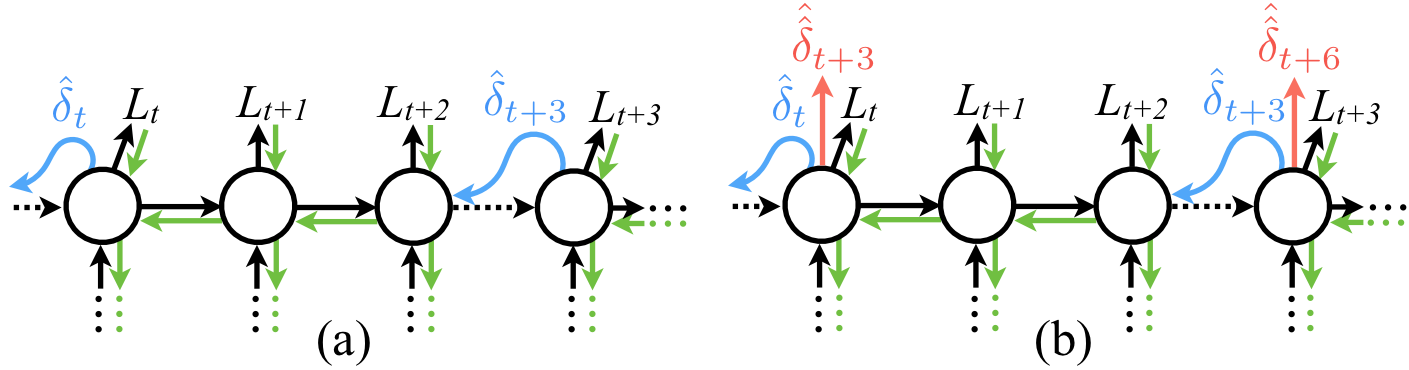}
\end{tabular}
\caption{\small (a) An RNN trained with truncated BPTT using DNI to communicate over time: Every timestep a recurrent core takes input and produces a hidden state $h_t$ and output $y_t$ which affects a loss $L_t$. The core is unrolled for $T$ steps (in this figure $T=3$). Gradients cannot propagate across the boundaries of BPTT, which limits the time dependency the RNN can learn to model. However, the recurrent core includes a synthetic gradient model which produces synthetic gradients $\hat{\delta}_t$ which can be used at the boundaries of BPTT to enable the last set of unrolled cores to communicate with the future ones. (b) In addition, as an auxiliary task, the network can also be asked to do future synthetic gradient prediction: an extra output $\hat{\hat{\delta}}_{t+T}$ is computed every timestep, and is trained to minimise $\|\hat{\hat{\delta}}_{t+T}-\hat{\delta}_{t+T}\|$.}
\label{fig:rnn}
\end{figure}


We begin by describing the high-level communication protocol that is used to allow asynchronously learning agents to communicate. 

As shown in \figref{fig:comms},  Sender $A$ sends a message $h_{A}$ to Receiver $B$. $B$ has a model $M_{B}$ of the utility of the message $h_{A}$. $B$'s model of utility $M_{B}$ is used to predict the feedback: an error signal $\hat{\delta}_{A} = M_{B}(h_{A}, s_B, c)$ based on the message $h_{A}$, the current state of $B$, $s_B$, and potentially any other information, $c$, that this module is privy to during training such as the label or context. The feedback $\hat{\delta}_{A}$ is sent back to $A$ which allows $A$ to be updated immediately. In time, $B$ can fully evaluate the true utility $\delta_{A}$ of the message received from $A$, and so $B$'s utility model can be updated to fit the true utility, reducing the disparity between $\hat{\delta}_{A}$ and $\delta_{A}$. 

This protocol allows $A$ to send messages to $B$ in a way that $A$ and $B$ are \emph{update decoupled} -- $A$ does not have to wait for $B$ to evaluate the true utility before it can be updated -- and $A$ can still learn to send messages of high utility to $B$.

We can apply this protocol to neural networks communicating, resulting in what we call Decoupled Neural Interfaces (DNI). For neural networks, the feedback error signal $\hat{\delta}_{A}$ can take different forms, \eg~gradients can be used as the error signal to work with backpropagation, target messages as the error signal to work with target propagation, or even a value (cumulative discounted future reward) to incorporate into a reinforcement learning framework. However, as a clear and easily analysable set of first steps into this important and mostly unexplored domain, we concentrate our empirical study on differentiable networks trained with backpropagation and gradient-based updates. Therefore, we focus on producing error gradients as the feedback $\hat{\delta}_{A}$ which we dub \emph{synthetic gradients}. 

\paragraph{Notation} 

To facilitate our exposition, it's useful to introduce some notation. Without loss of generality, consider neural networks as a graph of function operations (a finite chain graph in the case of a feed-forward models, an infinite chain in the case of recurrent ones, and more generally a directed acyclic graph). The forward execution of the network graph has a natural ordering due to the input dependencies of each functional node. We denote the function corresponding to step $i$ in a graph execution as $f_i$ and the composition of functions (\ie~the forward graph) from step $i$ to step $j$ inclusive as $\mathcal{F}^j_i.$ We denote the loss associated with layer, $i$, of the chain as $L_i$.


\subsection{Synthetic Gradient for Recurrent Networks}\label{sec:rnn}

We begin by describing  how our method of using synthetic gradients applies in the case of recurrent networks; in some ways this is simpler to reason about than feed-forward networks or more general graphs. 

An RNN applied to infinite stream prediction can be viewed as an infinitely unrolled recurrent core module $f$ with parameters $\theta$, such that the forward graph is $\mathcal{F}^\infty_1 = (f_i)_{i=1}^\infty$ where $f_i = f~\forall i$ and the core module propagates an output $y_i$ and state $h_i$ based on some input $x_i$: $y_i,h_i=f_i(x_i, h_{i-1})$.

At a particular point in time $t$ we wish to minimise $\sum_{\tau=t}^\infty L_\tau$. Of course, one cannot compute an update of the form
$\theta \leftarrow \theta - \alpha \sum_{\tau=t}^\infty \frac{\partial L_\tau}{\partial \theta}$
due to the infinite future time dependency. Instead, generally one considers a tractable time horizon $T$
\begin{equation*}
\begin{split}
\theta - \alpha \sum_{\tau=t}^\infty \frac{\partial L_\tau}{\partial \theta}
&= \theta - \alpha (\sum_{\tau=t}^{t+T} \frac{\partial L_\tau}{\partial \theta} + (\sum_{\tau=T+1}^\infty \frac{\partial L_\tau}{\partial h_T})\frac{\partial h_T}{\partial \theta})\\
 &= \theta - \alpha (\sum_{\tau=t}^{t+T} \frac{\partial L_\tau}{\partial \theta} + \delta_T\frac{\partial h_T}{\partial \theta})
\end{split}
\end{equation*}
and as in truncated BPTT, calculates $\sum_{\tau=t}^{t+T} \frac{\partial L_\tau}{\partial \theta}$ with backpropagation and approximates the remaining terms, beyond $t+T$, by using $\delta_T = 0$. This limits the time horizon over which updates to $\theta$ can be learnt, effectively limiting the amount of temporal dependency an RNN can learn. The approximation that $\delta_T=0$ is clearly naive, and by using an appropriately \emph{learned} approximation we can hope to do better. Treating the connection between recurrent cores at time $t+T$ as a Decoupled Neural Interface we can approximate $\delta_T$, with $\hat{\delta}_T = M_T(h_T)$ -- a learned approximation of the future loss gradients -- as shown and described in \figref{fig:rnn}~(a).

This amounts to taking the infinitely unrolled RNN as the full neural network $\mathcal{F}^\infty_1$, and chunking it into an infinite number of sub-networks where the recurrent core is unrolled for $T$ steps, giving $\mathcal{F}^{t+T-1}_t$. Inserting DNI between two adjacent sub-networks $\mathcal{F}^{t+T-1}_t$ and  $\mathcal{F}^{t+2T-1}_{t+T}$ allows the recurrent network to learn to communicate to its future self, without being update locked to its future self. From the view of the synthetic gradient model, the RNN is predicting its own error gradients.

The synthetic gradient model $\hat{\delta}_T = M_T(h_T)$ is trained to predict the true gradients by minimising a distance $d(\hat{\delta}_T, \delta_T)$ to the target gradient $\delta_T$ -- in practice we find L$_2$ distance to work well. The target gradient is ideally the true gradient of future loss, $\sum_{\tau=T+1}^\infty \frac{\partial L_\tau}{\partial h_T}$, but as this is not a tractable target to obtain, we can use a target gradient that is itself bootstrapped from a synthetic gradient and then backpropagated and mixed with a number of steps of true gradient, \eg~$\delta_T = \sum_{\tau=T+1}^{2T} \frac{\partial L_\tau}{\partial h_T} + \hat{\delta}_{2T+1}\frac{\partial h_{2T}}{\partial h_T}$. This bootstrapping is exactly analogous to bootstrapping value functions in reinforcement learning and allows temporal credit assignment to propagate beyond the boundary of truncated BPTT.

This training scheme can be implemented very efficiently by exploiting the recurrent nature of the network, as shown in \figref{fig:rnnprocess} in the Supplementary Material. In \sref{sec:exprnn} we show results on sequence-to-sequence tasks and language modelling, where using synthetic gradients extends the time dependency the RNN can learn.

\paragraph{Auxiliary Tasks}
We also propose an extension to aid learning of synthetic gradient models for RNNs, which is to introduce another auxiliary task from the RNN, described in \figref{fig:rnn}~(b). This extra prediction problem is designed to promote coupling over the maximum time span possible, requiring the recurrent core to explicitly model short term and long term synthetic gradients, helping propagate gradient information backwards in time. This is also shown to further increase performance in \sref{sec:exprnn}.

\subsection{Synthetic Gradient for Feed-Forward Networks}\label{sec:ff}

\begin{figure}[t]
\centering
\begin{tabular}{cccc}
\includegraphics[width=0.45\textwidth]{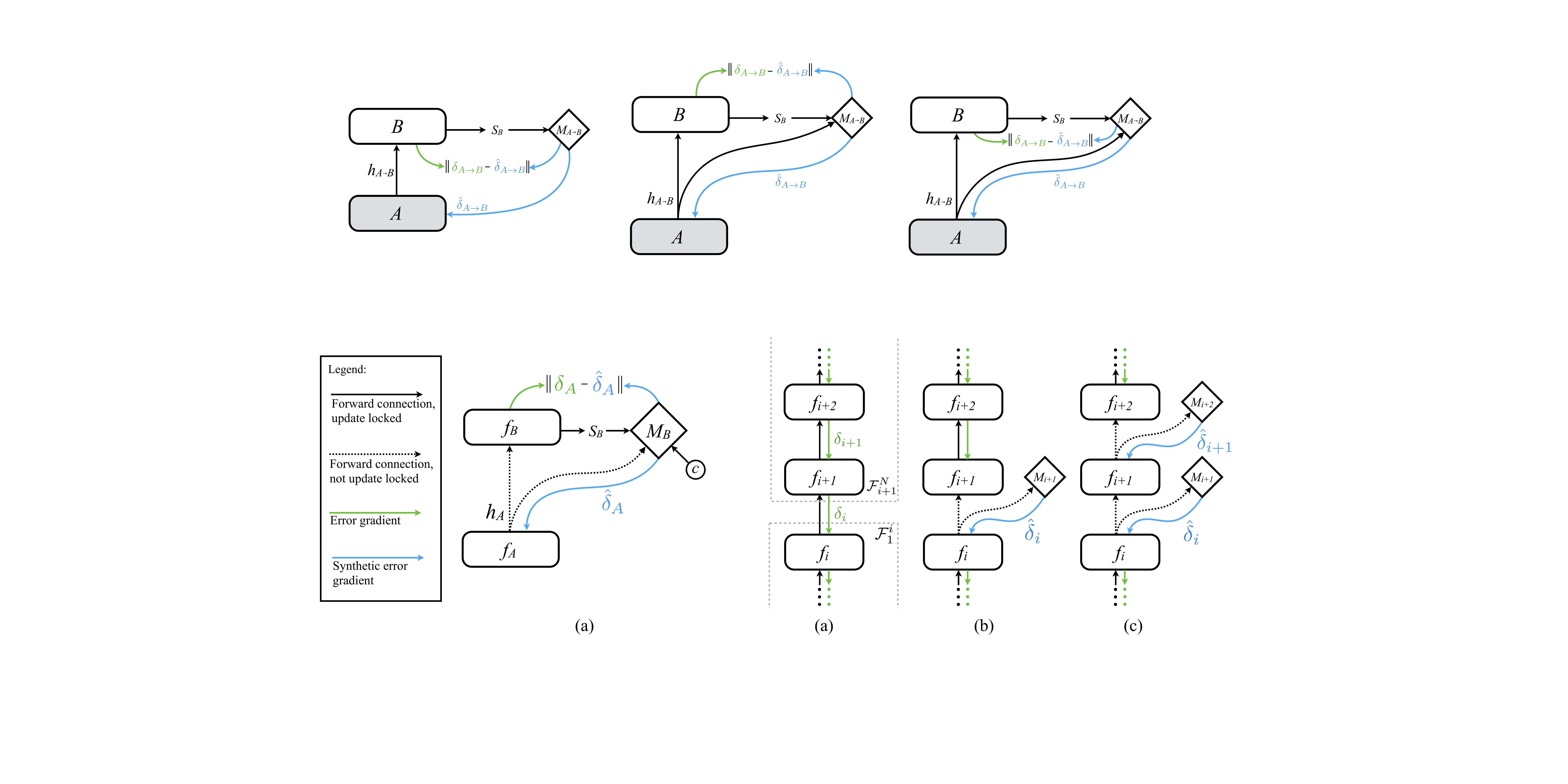}
\end{tabular}
\caption{\small (a) A section of a vanilla feed-forward neural network $\mathcal{F}^N_1$. (b) Incorporating one synthetic gradient model for the output of layer $i$. This results in two sub-networks $\mathcal{F}^i_1$ and $\mathcal{F}^N_{i+1}$ which can be updated independently. (c) Incorporating multiple synthetic gradient models after every layer results in $N$ independently updated layers.}
\label{fig:fftypes}
\end{figure}

As another illustration of DNIs, we now consider feed-forward networks consisting of $N$ layers $f_i, i \in \{1,\ldots,N\}$, each taking an input $h_{i-1}$ and producing an output $h_{i} = f_i(h_{i-1})$, where $h_0=x$ is the input data. The forward execution graph of the full network can be denoted as as $\mathcal{F}^N_1$, a section of which is illustrated in \figref{fig:fftypes}~(a).

Define the loss imposed on the output of the network as $L=L_N$. Each layer $f_i$ has parameters $\theta_i$ that can be trained jointly to minimise $L(h_N)$ with a gradient-based update rule
\begin{equation*}
\theta_i \leftarrow \theta_i - \alpha~\delta_i \frac{\partial h_i}{\partial \theta_i}~;~~\delta_i=\frac{\partial L}{\partial h_i}
\label{eqn:sgd}
\end{equation*}
where $\alpha$ is the learning rate and $\frac{\partial L}{\partial h_i}$ is computed with backpropagation. The reliance on  $\delta_i$ means that the update to layer $i$ can only occur after the remainder of the network, \ie~$\mathcal{F}^N_{i+1}$ (the sub-network of layers between layer $i+1$ and layer $N$ inclusive) has executed a full forward pass, generated the loss $L(h_N)$, then backpropagated the gradient through every successor layer in reverse order. Layer $i$ is therefore update locked to $\mathcal{F}^N_{i+1}$.

To remove the update locking of layer $i$ to $\mathcal{F}^N_{i+1}$ we can use the communication protocol described previously. Layer $i$ sends $h_i$ to layer $i+1$, which has a communication model $M_{i+1}$ that produces a synthetic error gradient $\hat{\delta}_i = M_{i+1}(h_i)$, as shown in \figref{fig:fftypes}~(b), which can be used immediately to update layer $i$ and all the other layers in $\mathcal{F}^i_1$
\begin{equation*}
\theta_n \leftarrow \theta_n - \alpha~\hat{\delta}_i \frac{\partial h_i}{\partial \theta_n},~~n \in \{1,\ldots,i\}.
\label{eqn:sgdbprop}
\end{equation*}
To train the parameters of the synthetic gradient model $M_{i+1}$, we simply wait for the true error gradient $\delta_i$ to be computed (after a full forwards and backwards execution of $\mathcal{F}^N_{i+1}$), and fit the synthetic gradient to the true gradients by minimising $\|\hat{\delta}_i - \delta_i\|^2_2$.

Furthermore, for a feed-forward network, we can use synthetic gradients as communication feedback to decouple every layer in the network, as shown in \figref{fig:fftypes}~(c). The execution of this process is illustrated in \figref{fig:ffprocess} in the Supplementary Material. In this case, since the target error gradient $\delta_i$ is produced by backpropagating $\hat{\delta}_{i+1}$ through layer $i+1$, $\delta_i$ is not the true error gradient, but an estimate bootstrapped from synthetic gradient models later in the network. Surprisingly, this does not cause errors to compound and learning remains stable even with many layers, as shown in \sref{sec:expff}. 

Additionally, if any supervision or context $c$ is available at the time of synthetic gradient computation, the synthetic gradient model can take this as an extra input, $\hat{\delta}_i = M_{i+1}(h_i, c)$.

This process allows a layer to be updated as soon as a forward pass of that layer has been executed. This paves the way for sub-parts or layers of networks to be trained in an asynchronous manner, something we show in \sref{sec:expff}.

\subsection{Arbitrary Network Graphs} 
Although we have explicitly described the application of DNIs for communication between layers in feed-forward networks, and between recurrent cores in recurrent networks, there is nothing to restrict the use of DNIs for arbitrary network graphs. The same procedure can be applied to any network or collection of networks, any number of times. An example is in \sref{sec:expmulti} where we show communication between two RNNs, which tick at different rates, where the communication can be learnt by using synthetic gradients.

\subsection{Mixing Real \& Synthetic Gradients}
In this paper we focus on the use of synthetic gradients to replace real backpropagated gradients in order to achieve update unlocking. However, synthetic gradients could also be used to augment real gradients. Mixing real and synthetic gradients results in $BP(\lambda)$, an algorithm anolgous to $TD(\lambda)$ for reinforcement learning~\citep{Sutton98}. This can be seen as a generalized view of synthetic gradients, with the algorithms given in this section for update unlocked RNNs and feed-forward networks being specific instantiations of $BP(\lambda)$. This generalised view is discussed further in \sref{sec:unified} in the Supplementary Material.

\section{Experiments}\label{sec:exp}

In this section we perform empirical expositions of the use of DNIs and synthetic gradients, first by applying them to RNNs in \sref{sec:exprnn} showing that synthetic gradients extend the temporal correlations an RNN can learn. Secondly, in \sref{sec:expmulti} we show how a hierarchical, two-timescale system of networks can be jointly trained using synthetic gradients to propagate error signals between networks. Finally, we demonstrate the ability of DNIs to allow asynchronous updating of layers a feed-forward network in \sref{sec:expff}. More experiments can be found in \sref{sec:appexp} in the Supplementary Material.

\subsection{Recurrent Neural Networks}\label{sec:exprnn}

\begin{table*}[t]
\begin{center}\small
\setlength{\tabcolsep}{3pt}
\begin{tabular}{r|ccccccc|ccccc|ccccc}
\toprule
         & \multicolumn{7}{c|}{BPTT}                     & \multicolumn{5}{c|}{DNI} & \multicolumn{5}{c}{DNI + Aux}                     \\
 $T=$ &  2  &  3  & 4  &  5 & 8 & 20 & 40  &  2  &  3  &  4  &  5 & 8 & 2 & 3 & 4 & 5 & 8   \\
\midrule
Copy & 7 & 8 & 10 & 8 & - & - & - & 16 & 14 & 18 & 18 & - & 16 & 17 & 19 & 18 & - \\

Repeat Copy & 7 & 5 & 19 & 23 & - & - & - & 39 & 33 & 39 & 59 & - &39 & 59 & 67 & 59 & - \\
      
Penn Treebank & 1.39 & 1.38 & 1.37 & 1.37 & 1.35 & 1.35 & 1.34 &  1.37 & 1.36 & 1.35 & 1.35 & 1.34 & 1.37 & 1.36 & 1.35 & 1.35 & 1.33 \\
\bottomrule
\end{tabular}
\end{center}
\caption{\small Results for applying DNI to RNNs. Copy and Repeat Copy task performance is reported as the maximum sequence length that was successfully modelled (higher is better), and Penn Treebank results are reported in terms of test set bits per character (lower is better) at the point of lowest validation error. No learning rate decreases were performed during training.}
\label{table:rnnresults}
\end{table*}

\begin{figure*}[t]
\centering
\begin{tabular}{cccc}
\includegraphics[width=1.0\textwidth]{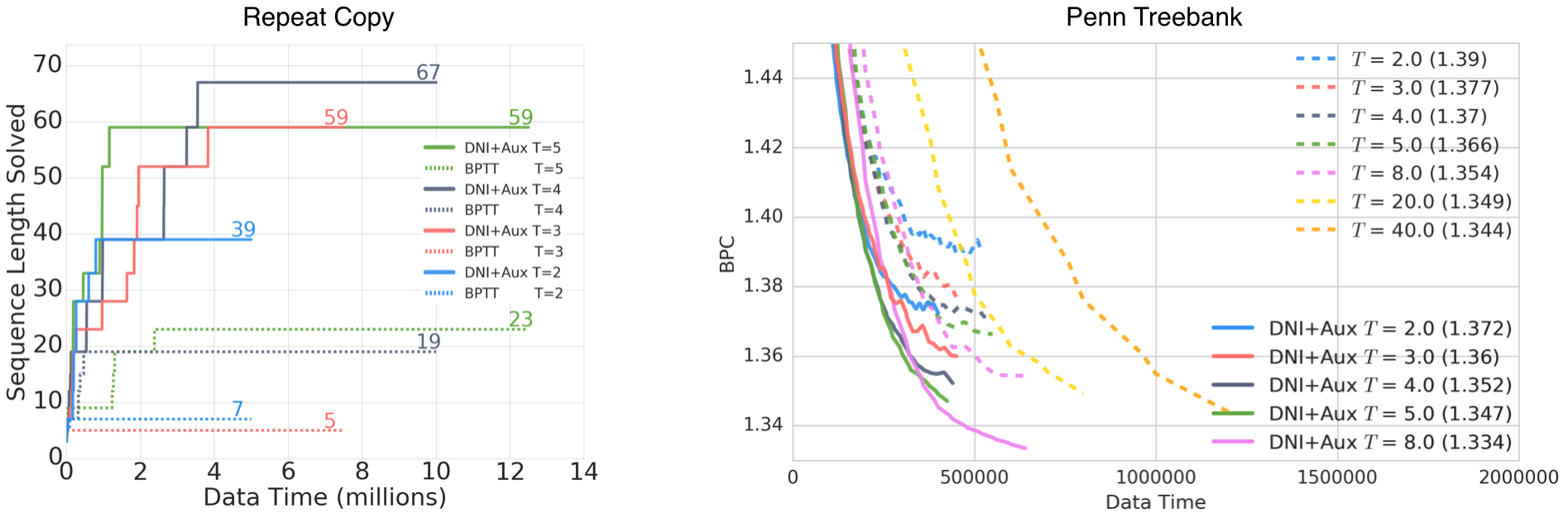}
\end{tabular}
\caption{\small \emph{Left:} The task progression during training for the Repeat Copy task. All models were trained for 2.5M iterations, but the varying unroll length $T$ results in different quantities of data consumed. The x-axis shows the number of samples consumed by the model, and the y-axis the time dependency level solved by the model -- step changes in the time dependency indicate that a particular time dependency is deemed solved. DNI+Aux refers to DNI with the additional future synthetic gradient prediction auxiliary task. \emph{Right:} Test error in bits per character (BPC) for Penn Treebank character modelling. We train the RNNs with different BPTT unroll lengths with DNI (solid lines) and without DNI (dashed lines). Early stopping is performed based on the validation set. Bracketed numbers give final test set BPC.}
\label{fig:rnngraphs}
\end{figure*}

Here we show the application of DNIs to recurrent neural networks as discussed in \sref{sec:rnn}. We test our models on the Copy task, Repeat Copy task, as well as character-level language modelling.

For all experiments we use an LSTM~\citep{Hochreiter97} of the form in~\citep{Graves13}, whose output is used for the task at hand, and additionally as input to the synthetic gradient model (which is shared over all timesteps). The LSTM is unrolled for $T$ timesteps after which backpropagation through time (BPTT) is performed. We also look at incorporating an auxiliary task which predicts the output of the synthetic gradient model $T$ steps in the future as explained in \sref{sec:rnn}. The implementation details of the RNN models are given in \sref{sec:apprnndetails} in the Supplementary Material. 

\paragraph{Copy and Repeat Copy}
We first look at two synthetic tasks -- Copy and Repeat Copy tasks from~\citep{Graves14}. Copy involves reading in a sequence of $N$ characters and after a stop character is encountered, must repeat the sequence of $N$ characters in order and produce a final stop character. Repeat Copy must also read a sequence of $N$ characters, but after the stop character, reads the number, $R$, which indicates the number of times it is required to copy the sequence, before outputting a final stop character. Each sequence of reading and copying is an episode, of length $T_\text{task}=N+3$ for Copy and $T_\text{task}=NR + 3$ for Repeat Copy.

While normally the RNN would be unrolled for the length of the episode before BPTT is performed, $T=T_\text{task}$, we wish to test the length of time the RNN is able to model with and without DNI bridging the BPTT limit. We therefore train the RNN with truncated BPTT: $T\in \{2,3,4,5 \}$ with and without DNI, where the RNN is applied continuously and across episode boundaries. For each problem, once the RNN has solved a task with a particular episode length (averaging below 0.15 bits error), the task is made harder by extending $N$ for Copy and Repeat Copy, and also $R$ for Repeat Copy. 

\tblref{table:rnnresults} gives the results by reporting the largest $T_\text{task}$ that is successfully solved by the model. The RNNs without DNI generally perform as expected, with longer BPTT resulting in being able to model longer time dependencies. However, by introducing DNI we can extend the time dependency that is able to be modelled by an RNN. The additional computational complexity is negligible but we require an additional recurrent core to be stored in memory (this is illustrated in \figref{fig:rnnprocess} in the Supplementary Material). Because we can model larger time dependencies with a smaller $T$, our models become more data-efficient, learning faster and having to see less data samples to solve a task. Furthermore, when we include the extra task of predicting the synthetic gradient that will be produced $T$ steps in the future (DNI + Aux), the RNNs with DNI are able to model even larger time dependencies. For example with $T=3$ (\ie~performing BPTT across only three timesteps) on the Repeat Copy task, the DNI enabled RNN goes from being able to model 33 timesteps to 59 timesteps when using future synthetic gradient prediction as well. This is in contrast to without using DNI at all, where the RNN can only model 5 timesteps.  

\paragraph{Language Modelling}
We also applied our DNI-enabled RNNs to the task of character-level language modelling, using the Penn Treebank dataset~\citep{Marcus93}. We use an LSTM with 1024 units, which at every timestep reads a character and must predict the next character in the sequence. We train with BPTT with and without DNI, as well as when using future synthetic gradient prediction (DNI + Aux), with $T\in \{2,3,4,5,8 \}$ as well as strong baselines with $T=20,40$. We measure error in bits per character (BPC) as in~\citep{Graves13}, perform early stopping based on validation set error, and for simplicity do not perform any learning rate decay. For full experimental details please refer to \sref{sec:apprnndetails} in the Supplementary Material.

The results are given in \tblref{table:rnnresults}. Interestingly, with BPTT over only two timesteps ($T=2$) an LSTM can get surprisingly good accuracy at next character prediction. As expected, increasing $T$ results in increased accuracy of prediction. When adding DNI, we see an increase in speed of learning (learning curves can be found in \figref{fig:rnngraphs}~(Right) and \figref{fig:ptb} in the Supplementary Material), and models reaching greater accuracy (lower BPC) than their counterparts without DNI. As seen with the Copy and Repeat Copy task, future synthetic gradient prediction further increases the ability of the LSTM to model long range temporal dependencies -- an LSTM unrolled 5 timesteps with DNI and future synthetic gradient prediction gives the same BPC as a vanilla LSTM unrolled 20 steps, only needs 58\% of the data and is $2\times$ faster in wall clock time to reach 1.35BPC.

Although we report results only with LSTMs, we have found DNI to work similarly for vanilla RNNs and Leaky RNNs~\citep{Ollivier15}. 

\subsection{Multi-Network System}\label{sec:expmulti}

\begin{figure*}[t]
\centering
\begin{tabular}{cccc}
\includegraphics[width=0.45\textwidth]{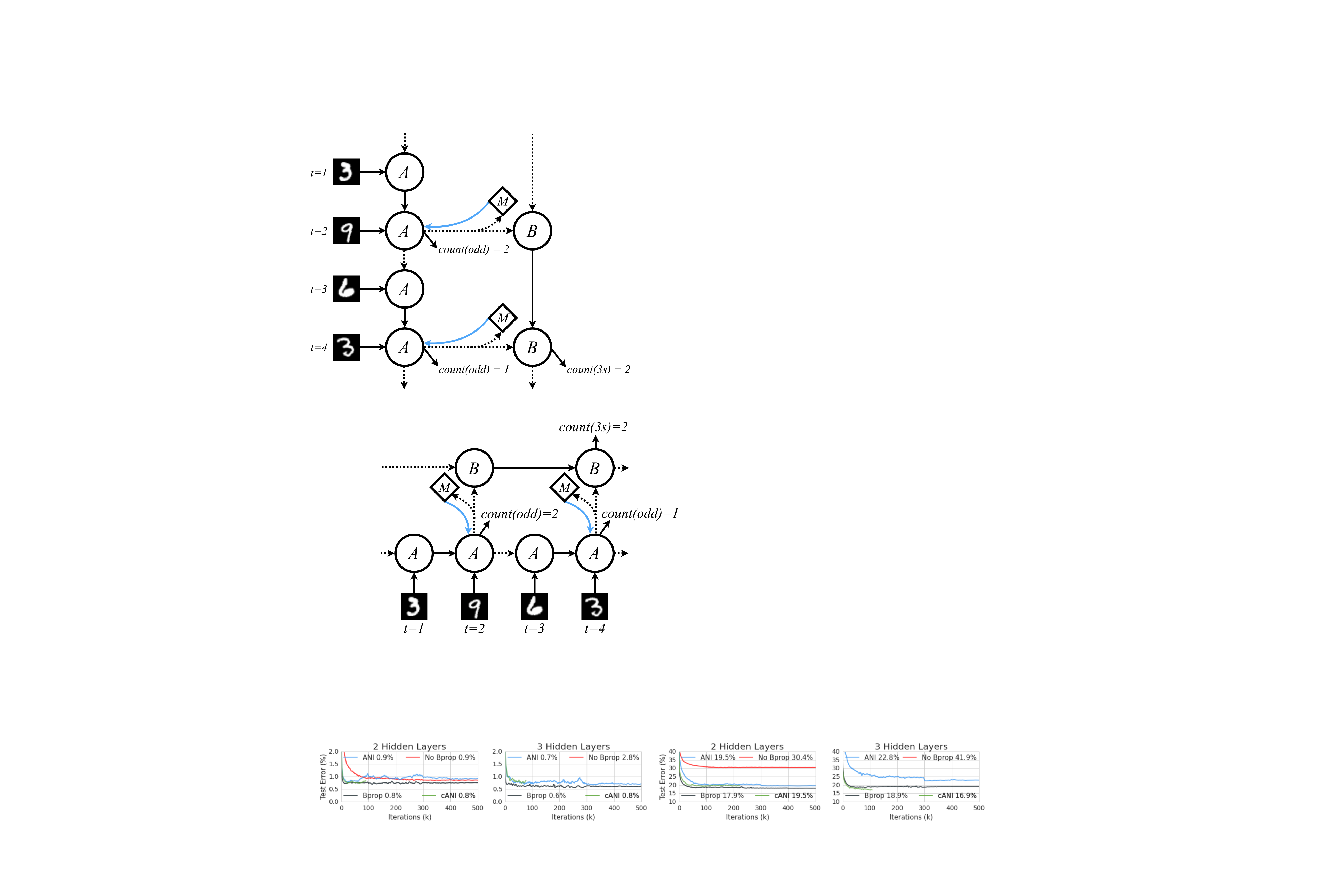}&
\hspace{-0.5em}\includegraphics[width=0.45\textwidth]{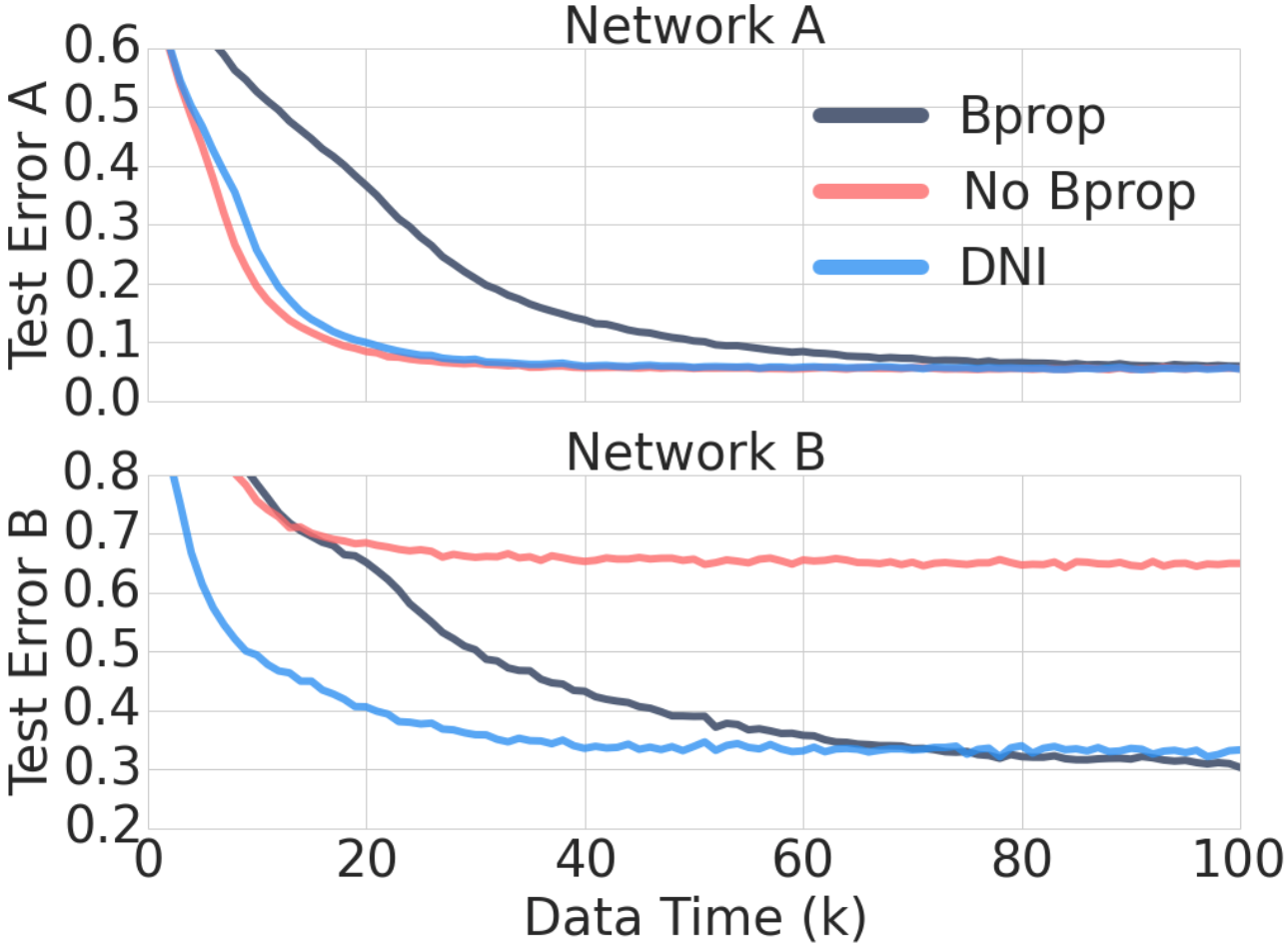}\\
\small(a)&\small(b)\\
\end{tabular}
\caption{\small (a) System of two RNNs communicating with DNI. Network A sees a datastream of MNIST digits and every $T$ steps must output the number of odd digits seen. Network B runs every $T$ steps, takes a message from Network A as input and must output the number of 3s seen over the last $T^2$ timesteps. Here is a depiction where $T=2$. (b) The test error over the course of training Network A and Network B with $T=4$. Grey shows when the two-network system is treated as a single graph and trained with backpropagation end-to-end, with an update every $T^2$ timesteps. The blue curves are trained where Network A and Network B are decoupled, with DNI (blue) and without DNI (red). When not decoupled (grey), Network A can only be updated every $T^2$ steps as it is update locked to Network B, so trains slower than if the networks are decoupled (blue and red). Without using DNI (red), Network A receives no feedback from Network B as to how to process the data stream and send a message, so Network B performs poorly. Using synthetic gradient feedback allows Network A to learn to communicate with Network B, resulting in similar final performance to the end-to-end learnt system (results remain stable after 100k steps).}
\label{fig:multinet}
\end{figure*}

In this section, we explore the use of DNI for communication between arbitrary graphs of networks. As a simple proof-of-concept, we look at a system of two RNNs, Network A and Network B, where Network B is executed at a slower rate than Network A, and must use communication from Network A to complete its task. The experimental setup is illustrated and described in \figref{fig:multinet}~(a). Full experimental details can be found in \sref{sec:appmultidetails} in the Supplementary Material.

First, we test this system trained end-to-end, with full backpropagation through all connections, which requires the joint Network A-Network B system to be unrolled for $T^2$ timesteps before a single weight update to both Network A and Network B, as the communication between Network A to Network B causes Network A to be update locked to Network B. We the train the same system but using synthetic gradients to create a learnable bridge between Network A and Network B, thus decoupling Network A from Network B. This allows Network A to be updated $T$ times more frequently, by using synthetic gradients in place of true gradients from Network B. 

\figref{fig:multinet}~(b) shows the results for $T=4$. Looking at the test error during learning of Network A (\figref{fig:multinet}~(b) Top), it is clear that being decoupled and therefore updated more frequently allows Network A to learn much quicker than when being locked to Network B, reaching final performance in under half the number of steps. Network B also trains faster with DNI (most likely due to the increased speed in learning of Network A), and reaches a similar final accuracy as with full backpropagation (\figref{fig:multinet}~(b) Bottom). When the networks are decoupled but DNI is not used (\ie~no gradient is received by Network A from Network B), Network A receives no feedback from Network B, so cannot shape its representations and send a suitable message, meaning Network B cannot solve the problem.

\subsection{Feed-Forward Networks}\label{sec:expff}
In this section we apply DNIs to feed-forward networks in order to allow asynchronous or sporadic training of layers, as might be required in a distributed training setup. As explained in \sref{sec:ff}, making layers decoupled by introducing synthetic gradients allows the layers to communicate with each other without being update locked.


\begin{figure*}[t]
 \centering
 \begin{tabular}{cccc}
 \includegraphics[width=0.7\textwidth]{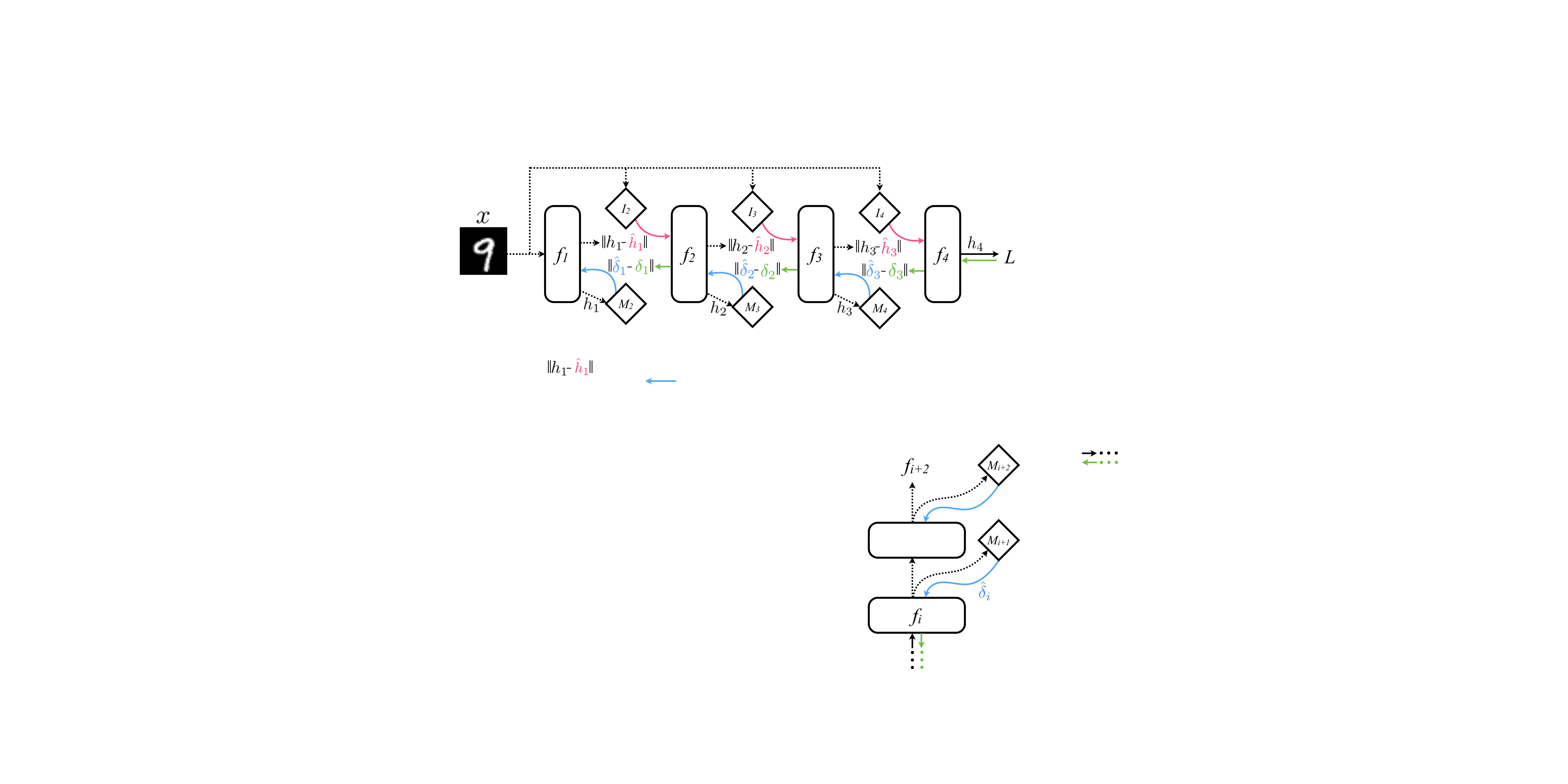}
 \end{tabular}
 \caption{\small Completely unlocked feed-forward network training allowing forward and update decoupling of layers.}
 \label{fig:drasticnet}
 \end{figure*}

\begin{figure*}[t]
\centering
\begin{tabular}{c}
\includegraphics[width=1.0\textwidth]{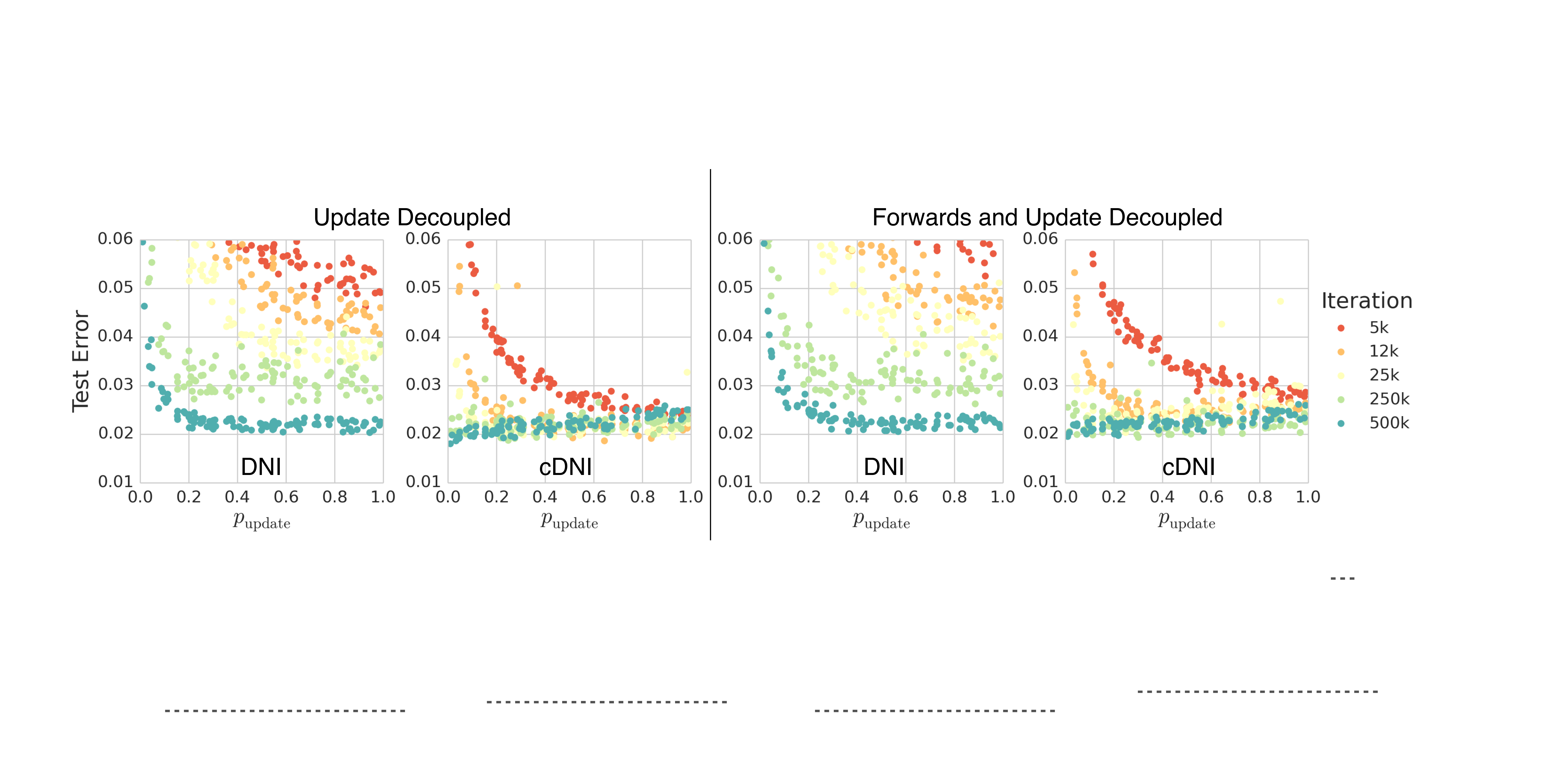}
\end{tabular}
\caption{\small \emph{ Left:} Four layer FCNs trained on MNIST using DNI between every layer, however each layer is trained stochastically -- after every forward pass, a layer only does a backwards pass with probability $p_{\text{update}}$. Population test errors are shown after different numbers of iterations (turquoise is at the end of training after 500k iterations). The purple diamond shows the result when performing regular backpropagation, requiring a synchronous backwards pass and therefore $p_{\text{update}}=1$. When using cDNIs however, with only 5\% probability of a layer being updated the network can train effectively. \emph{Right:} The same setup as previously described however we also use a synthetic \emph{input} model before every layer, which allows the network to also be \emph{forwards decoupled}. Now every layer is trained completely asynchronously, where with probability $1-p_{\text{update}}$ a layer does not do a forward pass or backwards pass -- effectively the layer is ``busy'' and cannot be touched at all.}
\label{fig:drastic}
\end{figure*}

\paragraph{Asynchronous Updates}
To demonstrate the gains by decoupling layers given by DNI, we perform an experiment on a four layer FCN model on MNIST, where the backwards pass and update for every layer occurs in random order and only with some probability $p_{\text{update}}$ (\ie~a layer is only updated after its forward pass $p_{\text{update}}$ of the time). This completely breaks backpropagation, as for example the first layer would only receive error gradients with probability $p_{\text{update}}^{3}$ and even then, the system would be constrained to be synchronous. However, with DNI bridging the communication gap between each layer, the stochasticity of a layer's update does not mean the layer below cannot update, as it uses synthetic gradients rather than backpropagated gradients. We ran 100 experiments with different values of $p_{\text{update}}$ uniformly sampled between 0 and 1. The results are shown in \figref{fig:drastic}~(Left) for DNI with and without conditioning on the labels. With $p_{\text{update}}=0.2$ the network can still train to 2\% accuracy. Incredibly, when the DNI is conditioned on the labels of the data (a reasonable assumption if training in a distributed fashion), the network trains perfectly with only 5\% chance of an update, albeit just slower.

\paragraph{Complete Unlock}
As a drastic extension, we look at making feed-forward networks completely asynchronous, by removing forward locking as well. In this scenario, every layer has a synthetic gradient model, but also a synthetic \emph{input} model -- given the data, the synthetic input model produces an approximation of what the input to the layer will be. This is illustrated in \figref{fig:drasticnet}. Every layer can now be trained independently, with the synthetic gradient and input models trained to regress targets produced by neighbouring layers. The results on MNIST are shown in \figref{fig:drastic}~(Right), and at least in this simple scenario, the completely asynchronous collection of layers train independently, but co-learn to reach 2\% accuracy, only slightly slower. More details are given in the Supplementary Material.




\section{Discussion \& Conclusion}\label{sec:conclusion}


In this work we introduced a method, \emph{DNI using synthetic gradients}, which allows decoupled communication between components, such that they can be independently updated. 
We demonstrated significant gains from the increased time horizon that DNI-enabled RNNs are able to model, as well as faster convergence. 
We also demonstrated the application to a multi-network system: a communicating pair of fast- and slow-ticking RNNs can be decoupled, greatly accelarating learning.
Finally, we showed that the method can be used facilitate distributed training by enabling us to completely decouple all the layers of a feed-forward net -- thus allowing them to be trained asynchronously, non-sequentially, and sporadically.

It should be noted that while this paper introduces and shows empirical justification for the efficacy of DNIs and synthetic gradients, the work of \cite{wojtek17} delves deeper into the analysis and theoretical understanding of DNIs and synthetic gradients, confirming the convergence properties of these methods and modelling impacts of using synthetic gradients.

To our knowledge this is the first time that neural net modules have been decoupled, and the update locking has been broken. This important result opens up exciting avenues of exploration -- including improving the foundations laid out here, and application to modular, decoupled, and asynchronous model architectures.


%
%

\nocite{langley00}

\defaultbibliography{shortstrings,vgg_local,vgg_other,current}
\defaultbibliographystyle{icml2017}


\begin{appendices}

\newpage
\twocolumn[
\icmltitle{\textbf{Supplementary Material for} \\ Decoupled Neural Interfaces using Synthetic Gradients }
\vskip 0.3in
]

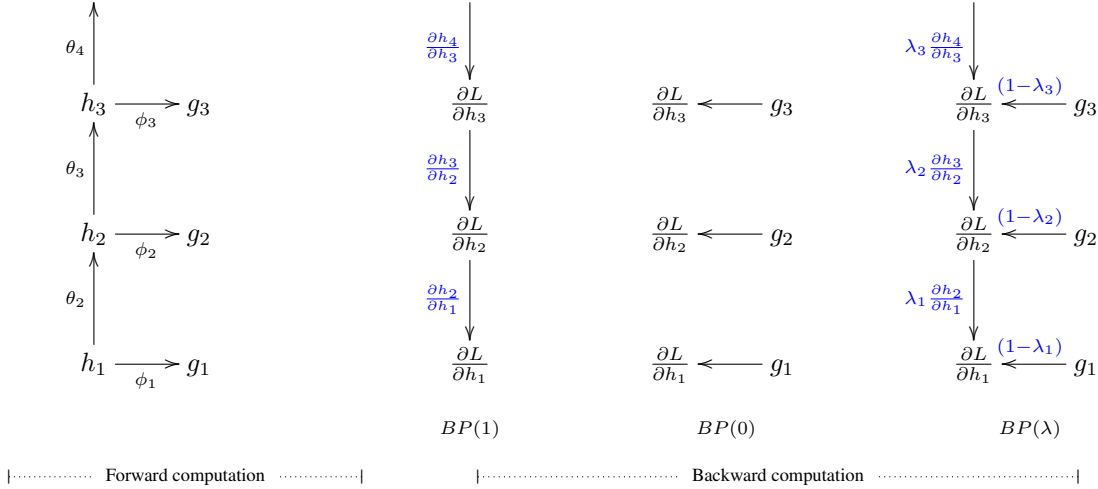
\begin{figure*}
    \xymatrix@R1pc@C2pc{
        &&
        		&&& \ar[dd]_{\blue{\dd{h_4}{h_3}}}
        		&& &
        		&& \ar[dd]_{\blue{\lambda_3 \dd{h_4}{h_3}}} & \\
\\
        & h_3 \ar[uu]^{\theta_4} \ar[r]_{\phi_3} & g_3
        		&&& \dd{L}{h_3} \ar[dd]_{\blue{\dd{h_3}{h_2}}}
        		&& \dd{L}{h_3} & g_3 \ar[l] 
        		&& \dd{L}{h_3} \ar[dd]_{\blue{\lambda_2 \dd{h_3}{h_2}}} & g_3 \ar[l]_{\blue{(1-\lambda_3)}} \\
\\
        & h_2 \ar[uu]^{\theta_3} \ar[r]_{\phi_2} & g_2
        		&&& \dd{L}{h_2} \ar[dd]_{\blue{\dd{h_2}{h_1}}}
        		&& \dd{L}{h_2} & g_2 \ar[l]
        		&& \dd{L}{h_2} \ar[dd]_{\blue{\lambda_1 \dd{h_2}{h_1}}} & g_2 \ar[l]_{\blue{(1-\lambda_2)}} \\
\\
        & h_1 \ar[uu]^{\theta_2} \ar[r]_{\phi_1} & g_1
        		&&& \dd{L}{h_1} 
        		&& \dd{L}{h_1} & g_1 \ar[l]
        		&& \dd{L}{h_1} & g_1 \ar[l]_{\blue{(1-\lambda_1)}} \\
         &&&&& \ar@{}[]|{BP(1)} && \ar@{}[r]|{BP(0)} &&& \ar@{}[r]|{BP(\lambda)} & \\
         \ar@{|.|}[rrrr]|{\text{\;\; Forward computation \;\;}} &&&&& \ar@{|.|}[rrrrrr]|{\text{\;\; Backward computation \;\;}} &&&&&&& \\
     } 
\caption{(Left) Forward computation of synthetic gradients. Arrows represent computations using parameters specified in label. (Right) Backward computation in BP($\lambda$). Each arrow may post-multiply its input by the specified value in blue label. BP(1) is equivalent to error backpropagation.
\label{fig:bplambda}
}
\end{figure*}

\section{Unified View of Synthetic Gradients}\label{sec:unified}

The main idea of this paper is to learn a \emph{synthetic gradient}, \ie~a separate prediction of the loss gradient for every layer of the network. The synthetic gradient can be used as a drop-in replacement for the backpropagated gradient. This provides a choice of two gradients at each layer: the gradient of the true loss, backpropagated from subsequent layers; or the synthetic gradient, estimated from the activations of that layer. 


In this section we present a unified algorithm, $BP(\lambda)$, that mixes these two gradient estimates as desired using a parameter $\lambda$. This allows the backpropagated gradient to be used insofar as it is available and trusted, but provides a meaningful alternative when it is not. This mixture of gradients is then backpropagated to the previous layer. 

\subsection{BP(0)}

We begin by defining our general setup and consider the simplest instance of synthetic gradients, $BP(0)$. We consider a feed-forward network with activations $h_k$ for $k\in \{1,\ldots,K\}$, and parameters $\theta_k$ corresponding to layers $k\in \{1,\ldots,K\}$. The goal is to optimize a loss function $L$ that depends on the final activations $h_K$. The key idea is to approximate the gradient of the loss, $g_k \approx \dd{L}{h_k}$, using a \emph{synthetic gradient}, $g_k$. The synthetic gradient is estimated from the activations at layer $k$, using a function $g_k = g(h_k, \phi_k)$ with parameters $\phi_k$. The overall loss can then be minimized by stochastic gradient descent on the synthetic gradient, 
\begin{align*}
\dd{L}{\theta_k} = \dd{L}{h_k} \dd{h_k}{\theta_k} \approx g_k \dd{h_k}{\theta_k}.
\end{align*}
In order for this approach to work, the synthetic gradient must also be trained to approximate the true loss gradient. Of course, it could be trained by regressing $g_k$ towards $\dd{L}{h_k}$, but our underlying assumption is that the backpropagated gradients are not available. Instead, we ``unroll'' our synthetic gradient just one step, 
\begin{align*}
g_k &\approx \dd{L}{h_k} = \dd{L}{h_{k+1}} \dd{h_{k+1}}{h_k} \approx g_{k+1} \dd{h_{k+1}}{h_k},
\end{align*}
and treat the unrolled synthetic gradient $z_k = g_{k+1} \dd{h_{k+1}}{h_k}$ as a constant training target for the synthetic gradient $g_k$. Specifically we update the synthetic gradient parameters $\phi_k$ so as to minimise the mean-squared error of these one-step unrolled training targets, by stochastic gradient descent on $\dd{(z_k - g_k)^2}{\phi_k}$. This idea is analogous to bootstrapping in the TD(0) algorithm for reinforcement learning \citep{sutton:td}. 

\subsection{BP($\lambda$)}

In the previous section we removed backpropagation altogether. We now consider how to combine synthetic gradients with a controlled amount of backpropagation. The idea of BP($\lambda$) is to mix together many different estimates of the loss gradient, each of which unrolls the chain rule for $n$ steps and then applies the synthetic gradient,
\begin{align*}
g^n_k &= g_{k+n} \dd{h_{k+n}}{h_{k+n-1}} ...  \dd{h_{k+1}}{h_k} \nonumber \\
&\approx \dd{L}{h_{k+n}} \dd{h_{k+n}}{h_{k+n-1}} ...  \dd{h_{k+1}}{h_k} \nonumber \\
&= \dd{L}{h_k}.
\end{align*}
We mix these estimators together recursively using a weighting parameter $\lambda_k$ (see Figure 1), 
\begin{align*}
\bar{g}_k &= \lambda_k \bar{g}_{k+1} \dd{h_{k+1}}{h_k} + (1 - \lambda_k) g_k.
\end{align*}
The resulting $\lambda$-weighted synthetic gradient $\bar{g}_k$ is a geometric mixture of the gradient estimates $g^1_k, ..., g^2_K$,
\begin{align*}
\bar{g}_k &= \sum_{n=k}^K c^n_k g^n_k.
\end{align*}
where $c^n_k = (1-\lambda_n) \prod_{j=k}^{n-1} \lambda_j$ is the weight of the $n$th gradient estimator $g^n_k$, and $c^K_k = 1 - \sum_{n=1}^{K-1} c^n_k$ is the weight for the final layer. This geometric mixture is analogous to the $\lambda$-return in $TD(\lambda)$ \citep{sutton:td}.

To update the network parameters $\theta$, we use the $\lambda$-weighted synthetic gradient estimate in place of the loss gradient,
\begin{align*}
\dd{L}{\theta_k} &= \dd{L}{h_k} \dd{h_k}{\theta_k} \approx \bar{g}_k \dd{h_k}{\theta_k}
\end{align*}
To update the synthetic gradient parameters $\phi_k$, we unroll the $\lambda$-weighted synthetic gradient by one step, $\bar{z}_k = \bar{g}_{k+1} \dd{h_{k+1}}{h_k}$, and treat this as a constant training target for the synthetic gradient $g_k$. Parameters are adjusted by stochastic gradient descent to minimise the mean-squared error between the synthetic gradient and its unrolled target, $\dd{(\bar{z}_k - g_k)^2}{\phi_k}$.

The two extreme cases of $BP(\lambda)$ result in simpler algorithms. If $\lambda_k=0~\forall k$ we recover the $BP(0)$ algorithm from the previous section, which performs no backpropagation whatsoever. If $\lambda_k=1~\forall k$ then the synthetic gradients are ignored altogether and we recover error backpropagation. For the experiments in this paper we have used binary values $\lambda_k \in \{0,1\}$. 

\subsection{Recurrent $BP(\lambda)$}

We now discuss how $BP(\lambda)$ may be applied to RNNs. We apply the same basic idea as before, using a synthetic gradient as a proxy for the gradient of the loss. However, network parameters $\theta$ and synthetic gradient parameters $\phi$ are now shared across all steps. There may also be a separate loss $l_k$ at every step $k$. The overall loss function is the sum of the step losses, $L = \sum_{k=1}^\infty l_k$. 

The synthetic gradient $g_k$ now estimates the cumulative loss from step $k+1$ onwards, $g_k \approx \dd{\sum_{j=k+1}^\infty l_j}{h_k}$. The $\lambda$-weighted synthetic gradient recursively combines these future estimates, and adds the immediate loss to provide an overall estimate of cumulative loss from step $k$ onwards,
\begin{align*}
\bar{g}_k &= \dd{l_k}{h_k} + \lambda_k \bar{g}_{k+1} \dd{h_{k+1}}{h_k} + (1 - \lambda_k) g_k.
\end{align*}
Network parameters are adjusted by gradient descent on the cumulative loss,
\begin{align*}
\dd{L}{\theta} &= \sum_{k=1}^\infty \dd{L}{h_k} \dd{h_k}{\theta} = \sum_{k=1}^\infty \dd{\sum_{j=k}^\infty l_j}{h_k} \dd{h_k}{\theta} \approx \sum_{k=1}^\infty \bar{g}_k \dd{h_k}{\theta}.
\end{align*}
To update the synthetic gradient parameters $\phi$, we again unroll the $\lambda$-weighted synthetic gradient by one step, $\bar{z}_k = \dd{l_k}{h_k} + \bar{g}_{k+1} \dd{h_{k+1}}{h_k}$, and minimise the MSE with respect to this target, over all time-steps, $\sum_{k=1}^\infty \dd{(\bar{z}_k - g_k)^2}{\phi}$.

We note that for the special case $BP(0)$, there is no backpropagation at all and therefore weights may be updated in a fully online manner. This is possible because the synthetic gradient estimates the gradient of cumulative future loss, rather than explicitly backpropagating the loss from the end of the sequence. 

Backpropagation-through-time requires computation from all time-steps to be retained in memory. As a result, RNNs are typically optimised in N-step chunks $[mN,(m+1)N]$. For each chunk $m$, the cumulative loss is initialised to zero at the final step $k=(m+1)N$, and then errors are backpropagated-through-time back to the initial step $k=mN$. However, this prevents the RNN from modelling longer term interactions. Instead, we can initialise the backpropagation at final step $k=(m+1)N$ with a synthetic gradient $g_k$ that estimates long-term future loss, and then backpropagate the synthetic gradient through the chunk. This algorithm is a special case of $BP(\lambda)$ where $\lambda_k=0$ if $k \mod N = 0$ and $\lambda_k=1$ otherwise. The experiments in \sref{sec:exprnn} illustrate this case.

\subsection{Scalar and Vector Critics}

One way to estimate the synthetic gradient is to first estimate the loss using a \emph{critic}, $v(h_k, \phi) \approx {\expect{L | h_k}}$, and then use the gradient of the critic as the synthetic gradient, $g_k = \dd{v(h_k, \phi)}{h_k} \approx \dd{L}{h_k}$. This provides a choice between a scalar approximation of the loss, or a vector approximation of the loss gradient, similar to the scalar and vector critics suggested by Fairbank \citep{fairbank:thesis}. 

These approaches have previously been used in control \citep{werbos:adp, fairbank:thesis} and model-based reinforcement learning \citep{heess:svg}. In these cases the dependence of total cost or reward on the policy parameters is computed by backpropagating through the trajectory. This may be viewed as a special case of the $BP(\lambda)$ algorithm; intermediate values of $\lambda<1$ were most successful in noisy environments \citep{heess:svg}.

It is also possible to use other styles of critics or error approximation techniques such as Feedback Alignment~\citep{lillicrap2016random}, Direct Feedback Alignment~\citep{NIPS2016_6441}, and Kickback~\citep{balduzzi2014kickback}) -- interestingly \cite{wojtek17} shows that they can all be framed in the synthetic gradients framework presented in this paper.

\section{Synthetic Gradients are Sufficient}\label{sec:sufficient}

In this section, we show that a function $f(h_t, \theta_{t+1:T})$, which depends only on the hidden activations $h_t$ and downstream parameters $\theta_{t+1:T}$, is sufficient to represent the gradient of a feedforward or recurrent network, without any other dependence on past or future inputs $x_{1:T}$ or targets $y_{1:T}$.

In (stochastic) gradient descent, parameters are updated according to (samples of) the expected loss gradient,
\begin{tiny}
\begin{align*}
\expectx{x_{1:T},y_{1:T}}{\dd{L}{\theta_t}} &= \expectx{x_{1:T},y_{1:T}}{\dd{L}{h_t} \dd{h_t}{\theta_t}} \\
&= \expectx{x_{1:T},y_{1:T}}{\expectx{x_{t+1:T},y_{t:T} | x_{1:t},y_{1:t-1}}{\dd{L}{h_t} \dd{h_t}{\theta_t}}} \\
&= \expectx{x_{1:T},y_{1:T}}{\expectx{x_{t+1:T},y_{t:T} | h_t }{\dd{L}{h_t}} \dd{h_t}{\theta_t}} \\
&= \expectx{x_{1:T},y_{1:T}}{g(h_t, \theta_{t+1:T}) \dd{h_t}{\theta_t}} \\
\end{align*}
\end{tiny}
where $g(h_t, \theta_{t+1:T}) = \expectx{x_{t+1:T},y_{t:T} | h_t }{\dd{L}{h_t}}$ is the expected loss gradient given hidden activations $h_t$. Parameters may be updated using samples of this gradient, $g(h_t, \theta_{t+1:T}) \dd{h_t}{\theta_t}$. 

The synthetic gradient $g(h_t, v_t) \approx g(h_t, \theta_{t+1:T})$ approximates this expected loss gradient at the current parameters $\theta_{t+1:T}$. If these parameters are frozen, then a sufficiently powerful synthetic gradient approximator can learn to perfectly represent the expected loss gradient. This is similar to an actor-critic architecture, where the neural network is the actor and the synthetic gradient is the critic.

In practice, we allow the parameters to change over the course of training, and therefore the synthetic gradient must learn online to track the gradient $g(h_t, \theta_{t+1:T})$


\begin{figure*}[t!]
 \centering
 \begin{tabular}{cccc}
 \includegraphics[width=0.8\textwidth]{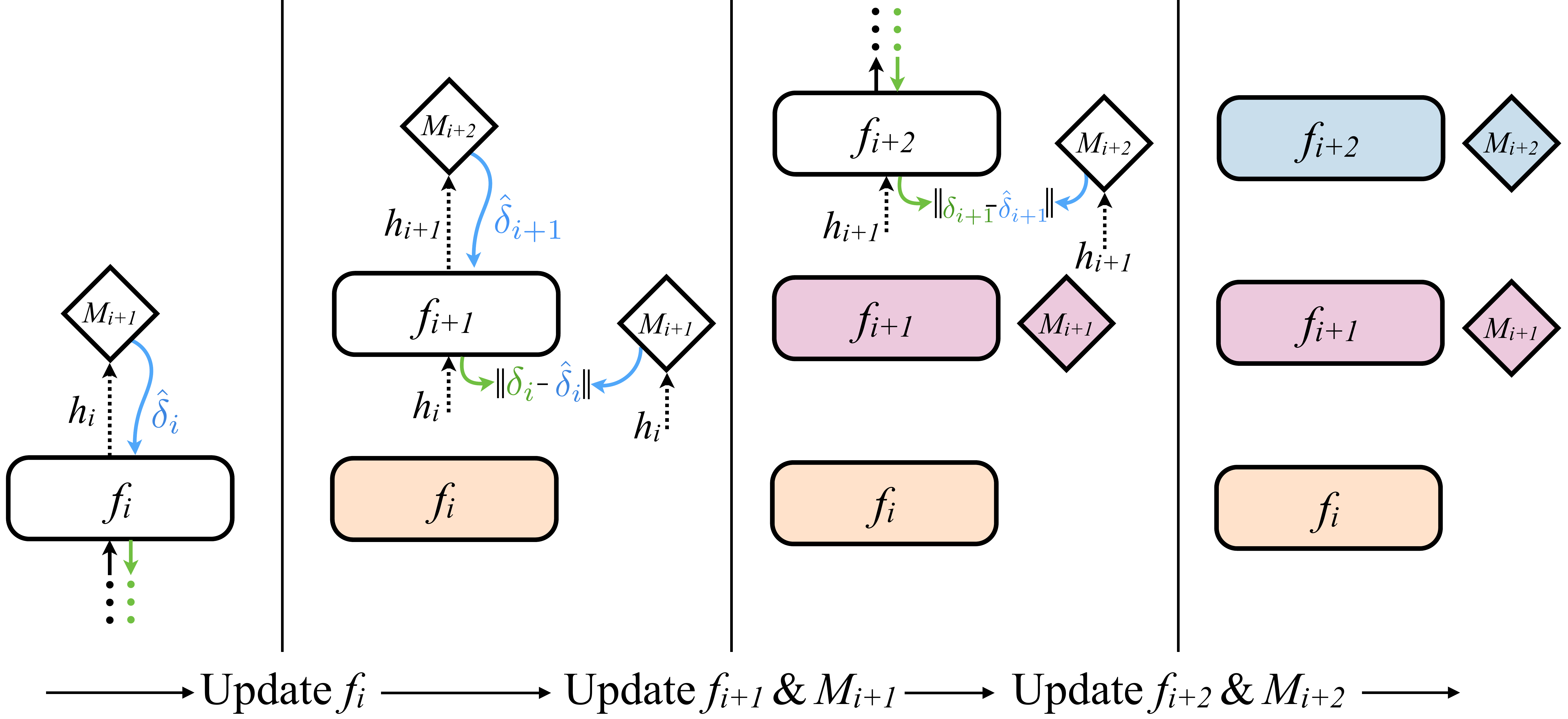}
 \end{tabular}
 \caption{\small The execution during training of a feed-forward network. Coloured modules are those that have been updated for this batch of inputs. First, layer $i$ executes it's forward phase, producing $h_i$, which can be used by $M_{i+1}$ to produce the synthetic gradient $\hat{\delta}_i$. The synthetic gradient is pushed backwards into layer $i$ so the parameters $\theta_i$ can be updated immediately. The same applies to layer $i+1$ where $h_{i+1}=f_{i+1}(h_i)$, and then $\hat{\delta}_{i+1}=M_{i+2}(h_{i+1})$ so layer $i+1$ can be updated. Next, $\hat{\delta}_{i+1}$ is backpropagated through layer $i+1$ to generate a target error gradient $\delta_i = f'_{i+1}(h_i)\hat{\delta}_{i+1}$ which is used as a target to regress $\hat{\delta}_i$ to, thus updating $M_{i+1}$. This process is repeated for every subsequent layer.}
 \label{fig:ffprocess}
 \end{figure*}
 
 \begin{figure*}[t]
 \centering
 \begin{tabular}{cccc}
 \includegraphics[width=0.7\textwidth]{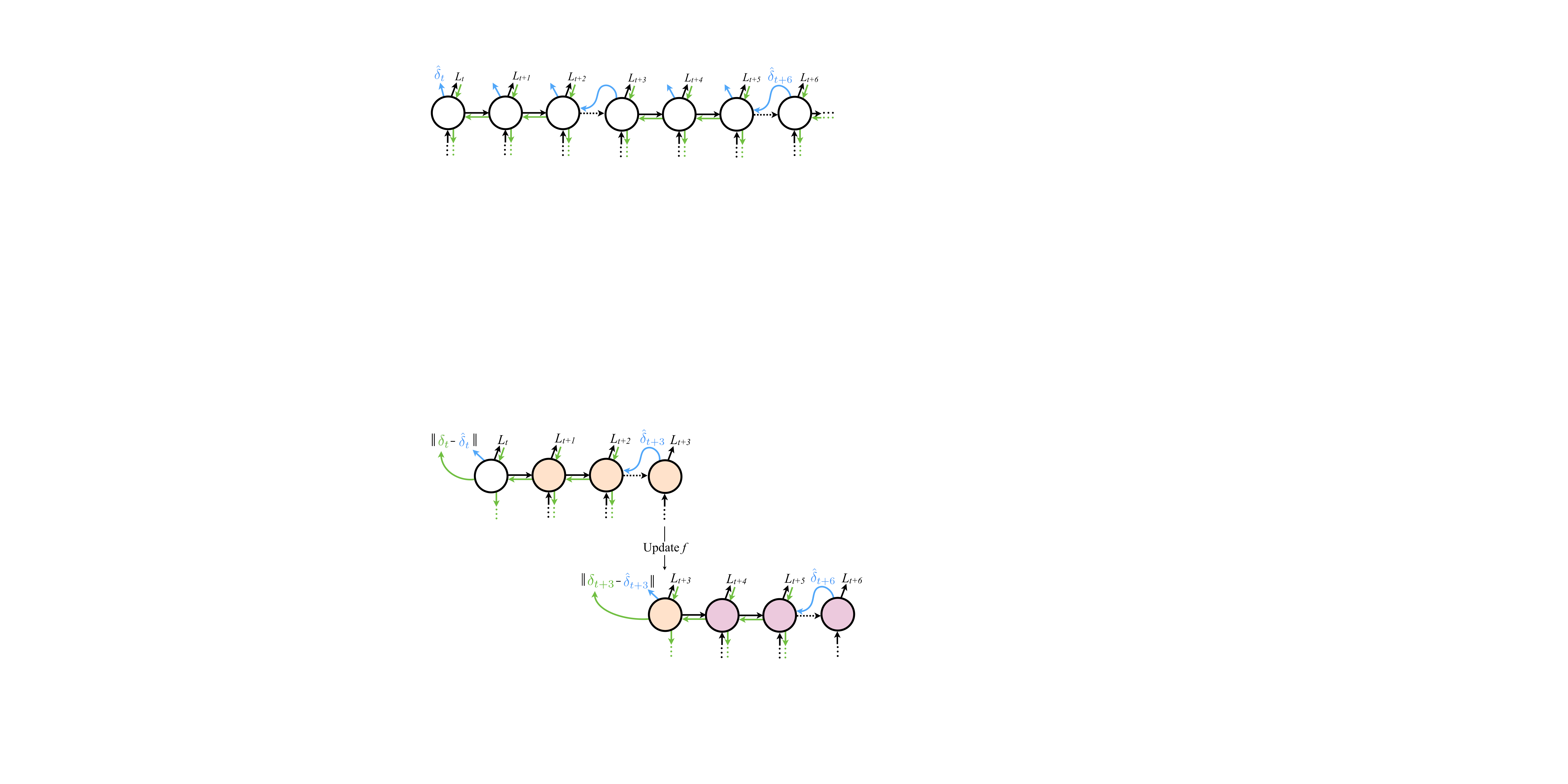}
 \end{tabular}
 \caption{\small The execution during training of an RNN, with a core function $f$, shown for $T=3$. Changes in colour indicate a weight update has occurred. The final core of the last unroll is kept in memory. Fresh cores are unrolled for $T$ steps, and the synthetic gradient from step $T$ (here $\hat{\delta}_{t+3}$ for example) is used to approximate the error gradient from the future. The error gradient is backpropagated through the earliest $T$ cores in memory, which gives a target error gradient for the last time a synthetic gradient was used. This is used to generate a loss for the synthetic gradient output of the RNN, and all the $T$ cores' gradients with respect to parameters can be accumulated and updated. The first $T$ cores in memory are deleted, and this process is repeated. This training requires an extra core to be stored in memory ($T+1$ rather than $T$ as in normal BPTT). Note that the target gradient of the hidden state that is regressed to by the synthetic gradient model is slightly stale, a similar consequence of online training as seen in RTRL~\citep{Williams89}.}
\label{fig:rnnprocess}
\end{figure*}

\section{Additional Experiments}\label{sec:appexp}

\begin{table}[t]
\begin{center}\small
\setlength{\tabcolsep}{4pt}
\begin{tabular}{lr|rrrr|rrrrr}
\toprule
      &         & \multicolumn{4}{c|}{MNIST (\% Error)}                     & \multicolumn{4}{c}{CIFAR-10 (\% Error)}                      \\
 \multicolumn{2}{r|}{Layers} &  \rotatebox[origin=c]{90}{~No Bprop}  &  \rotatebox[origin=c]{90}{Bprop}  &  \rotatebox[origin=c]{90}{DNI}  &  \rotatebox[origin=c]{90}{cDNI}  &  \rotatebox[origin=c]{90}{~No Bprop}  &  \rotatebox[origin=c]{90}{Bprop}  &  \rotatebox[origin=c]{90}{DNI}  &  \rotatebox[origin=c]{90}{cDNI}   \\
\midrule
\hline
\multirow{4}{*}{\rotatebox[origin=c]{90}{FCN}} &       3 &       9.3  &      2.0  &      1.9  &      2.2  &     54.9  &     43.5  &     42.5  &     48.5  \\
      &       4 &      12.6  &      1.8  &      2.2  &      1.9  &     57.2  &     43.0  &     45.0  &     45.1  \\
      &       5 &      16.2  &      1.8  &      3.4  &      1.7  &     59.6  &     41.7  &     46.9  &     43.5  \\
      &       6 &      21.4  &      1.8  &      4.3  &      1.6  &     61.9  &     42.0  &     49.7  &     46.8  \\
\midrule
\multirow{2}{*}{\rotatebox[origin=c]{90}{CNN}} &       3 &       0.9  &      0.8  &      0.9  &      1.0  &     28.7  &     17.9  &     19.5  &     19.0  \\
      &       4 &       2.8  &      0.6  &      0.7  &      0.8  &     38.1  &     15.7  &     19.5  &     16.4  \\
\bottomrule
\end{tabular}
\qquad\quad
\hspace{-3em}
\vtop{\vspace{0.5em}\hbox{\includegraphics[width=0.4\textwidth]{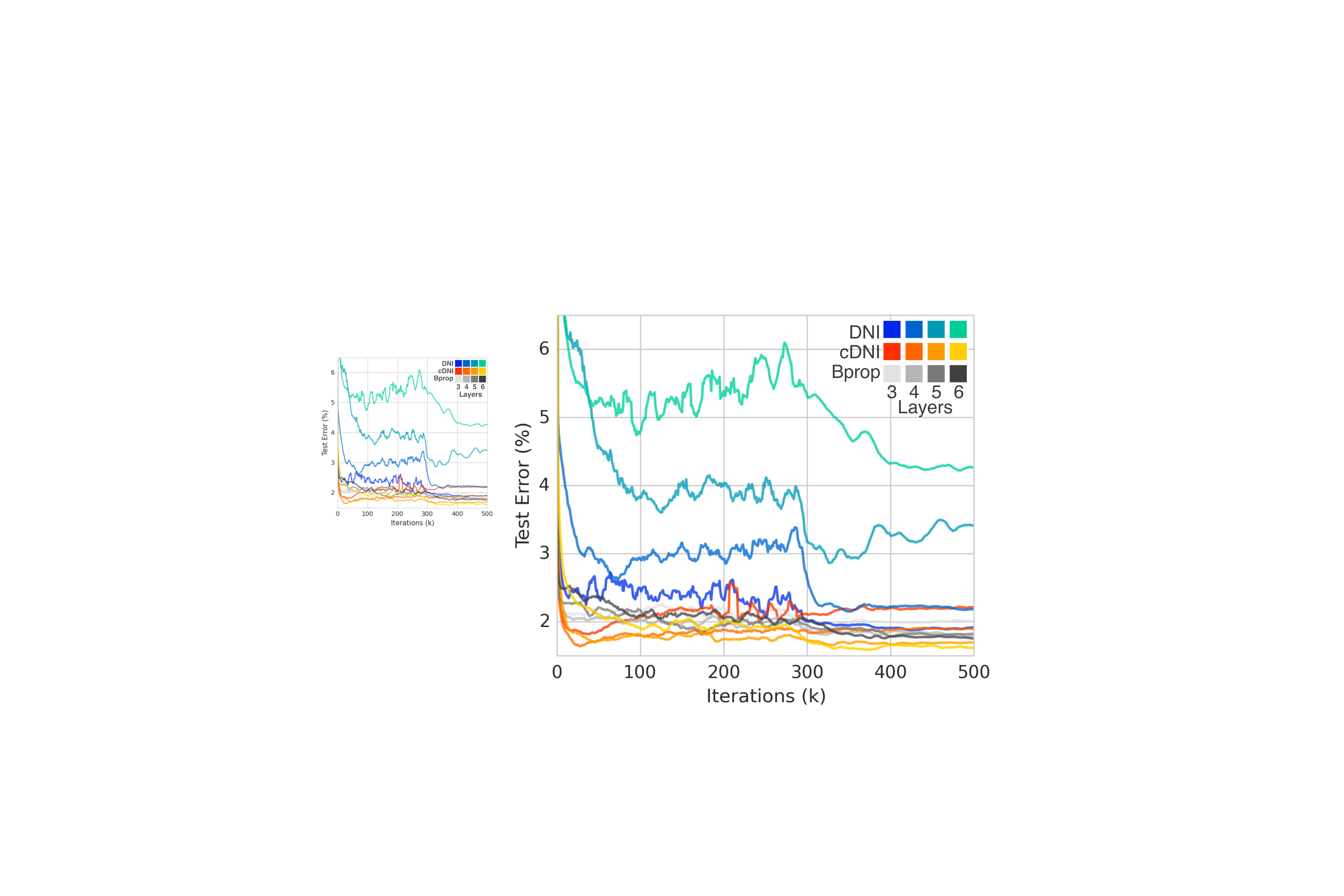}}}
\end{center}
\caption{\small Using DNI between every layer for FCNs and CNNs on MNIST and CIFAR-10. \emph{Left:} Summary of results, where values are final test error (\%) after 500k iterations. \emph{Right:} Test error during training of MNIST FCN models for regular backpropagation, DNI, and cDNI (DNI where the synthetic gradient model is also conditioned on the labels of the data).}
\label{table:ff}
\end{table}

\paragraph{Every layer DNI}
We first look at training an FCN for MNIST digit classification~\citep{lecun1998mnist}. For an FCN, ``layer'' refers to a linear transformation followed by batch-normalisation~\citep{Ioffe15} and a rectified linear non-linearity (ReLU)~\citep{Glorot11}. All hidden layers have the same number of units, 256. We use DNI as in the scenario illustrated in \figref{fig:fftypes}~(d), where DNIs are used between \emph{every} layer in the network. \Eg~for a four layer network (three hidden, one final classification) there will be three DNIs. In this scenario, every layer can be updated as soon as its activations have been computed and passed through the synthetic gradient model of the layer above, without waiting for any other layer to compute or loss to be generated. We perform experiments where we vary the depth of the model (between 3 and 6 layers), on MNIST digit classification and CIFAR-10 object recognition~\citep{Krizhevsky09}. Full implementation details can be found in \sref{sec:appffdetails}. 

Looking at the results in \tblref{table:ff} we can see that DNI does indeed work, successfully update-decoupling all layers at a small cost in accuracy, demonstrating that it is possible to produce effective gradients \emph{without either label or true gradient information}. Further, once we condition the synthetic gradients on the labels, we can successfully train deep models with very little degradation in accuracy. For example, on CIFAR-10 we can train a 5 layer model, with backpropagation achieving 42\% error, with DNI achieving 47\% error, and when conditioning the synthetic gradient on the label (cDNI) get 44\%. In fact, on MNIST we successfully trained up to 21 layer FCNs with cDNI to 2\% error (the same as with using backpropagation). Interestingly, the best results obtained with cDNI were with \emph{linear} synthetic gradient models.

As another baseline, we tried using historical, stale gradients with respect to activations, rather than synthetic gradients. We took an exponential average historical gradient, searching over the entire spectrum of decay rates and the best results attained on MNIST classification were 9.1\%, 11.8\%, 15.4\%, 19.0\% for 3 to 6 layer FCNs respectively -- marginally better than using zero gradients (no backpropagation) and far worse than the associated cDNI results of 2.2\%, 1.9\%, 1.7\%, 1.6\%. Note that the experiment described above used stale gradients with respect to the activations which do not correspond to the same input example used to compute the activation. In the case of a fixed training dataset, one could use the stale gradient from the same input, but it would be stale by an entire epoch and contains no new information so would fail to improve the model. Thus, we believe that DNI, which uses a parametric approximation to the gradient with respect to activations, is the most desirable approach.

This framework can be easily applied to CNNs~\citep{Lecun98}. The spatial resolution of activations from layers in a CNN results in high dimensional activations, so we use synthetic gradient models which themselves are CNNs without pooling and with resolution-preserving zero-padding. For the full details of the CNN models please refer to \sref{sec:appffdetails}. The results of CNN models for MNIST and CIFAR-10 are also found in \tblref{table:ff}, where DNI and cDNI CNNs perform exceptionally well compared to true backpropagated gradient trained models -- a three layer CNN on CIFAR-10 results in 17.9\% error with backpropagation, 19.5\% (DNI), and 19.0\% (cDNI).

\begin{figure*}[t]
\centering
\begin{tabular}{cccc}
\hspace{-4em}
\includegraphics[width=\textwidth]{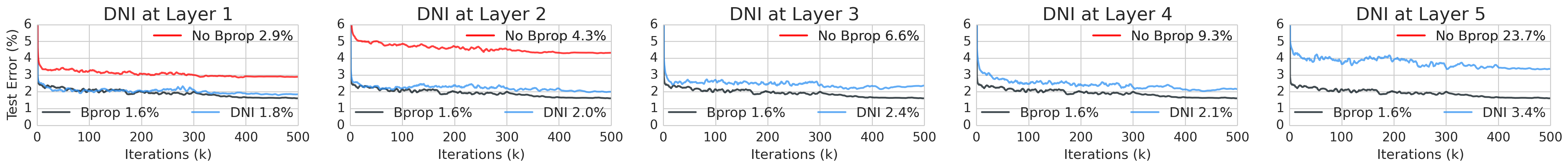}\\
\small (a)\\
\includegraphics[width=0.8\textwidth]{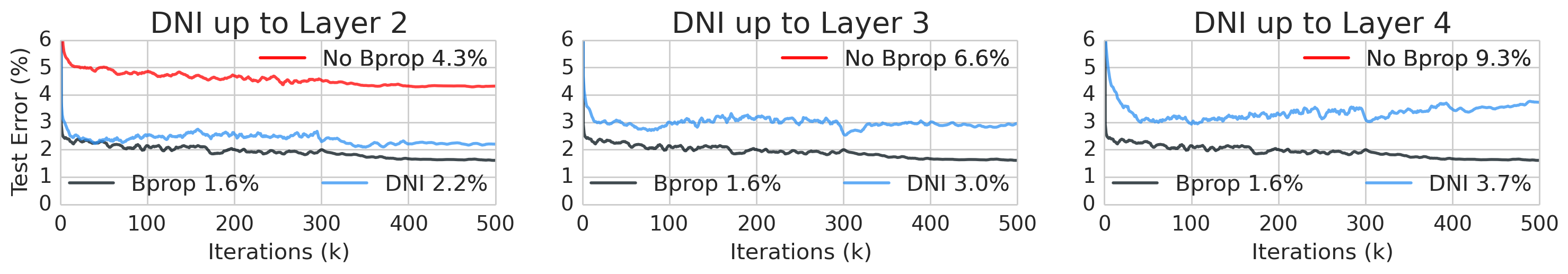}\\
\small (b)
\end{tabular}
\caption{\small Test error during training of a 6 layer fully-connected network on MNIST digit classification. Bprop (grey) indicates traditional, synchronous training with backpropagation, while DNI (blue) shows the use of a (a) single DNI used after a particular layer indicated above, and (b) every layer using DNI up to a particular depth. Without backpropagating any gradients through the connection approximated by DNI results in poor performance (red).}
\label{fig:fcnsingle}
\end{figure*}

\begin{figure*}[t]
\centering
\begin{tabular}{cccc}
\hspace{-4em}
\includegraphics[width=\textwidth]{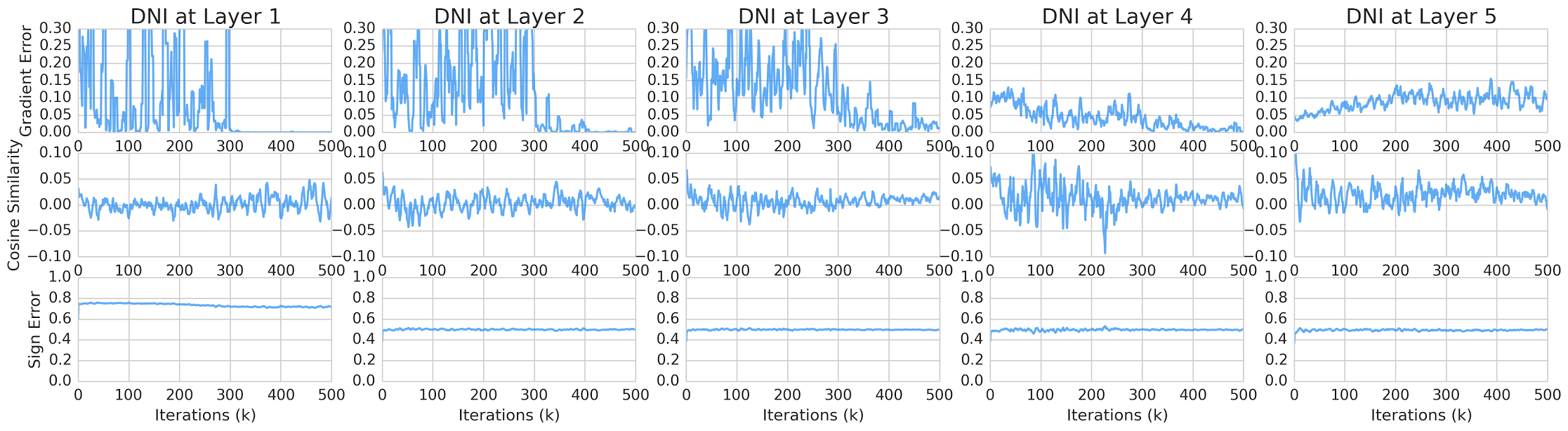}\\
\end{tabular}
\caption{\small Error between the synthetic gradient and the true backpropagated gradient for MNIST FCN where DNI is inserted at a single position. Sign error refers to the average number of dimensions of the synthetic gradient vector that do not have the same sign as the true gradient.}
\label{fig:graderr}
\end{figure*}

\paragraph{Single DNI}
We look at training an FCN for MNIST digit classification using a network with 6 layers (5 hidden layers, one classification layer), but splitting the network into two unlocked sub-networks by inserting a single DNI at a variable position, as illustrated in \figref{fig:fftypes}~(c).

\figref{fig:fcnsingle}~(a) shows the results of varying the depth at which the DNI is inserted. When training this 6 layer FCN with vanilla backpropagation we attain 1.6\% test error. Incorporating a single DNI between two layers results in between 1.8\% and 3.4\% error depending on whether the DNI is after the first layer or the penultimate layer respectively. If we decouple the layers without DNI, by just not backpropagating any gradient between them, this results in bad performance -- between 2.9\% and 23.7\% error for after layer 1 and layer 5 respectively. 

One can also see from \figref{fig:fcnsingle}~(a) that as the DNI module is positioned closer to the classification layer (going up in layer hierarchy), the effectiveness of it degrades. This is expected since now a larger portion of the whole system never observes true gradient. However, as we show in \sref{sec:expff}, using extra label information in the DNI module almost completely alleviates this problem.

We also plot the synthetic gradient regression error (L$_2$ distance), cosine distance, and the sign error (the number of times the sign of a gradient dimension is predicted incorrectly) compared to the true error gradient in \figref{fig:graderr}. Looking at the L$_2$ error, one can see that the error jumps initially as the layers start to train, and then the synthetic gradient model starts to fit the target gradients. The cosine similarity is on average very slightly positive, indicating that the direction of synthetic gradient is somewhat aligned with that of the target gradient, allowing the model to train. However, it is clear that the synthetic gradient is not tracking the true gradient very accurately, but this does not seem to impact the ability to train the classifiers.


\begin{figure*}[t]
\centering
\begin{tabular}{cccc}
\multicolumn{2}{c}{\small MNIST FCN}\\
\multicolumn{2}{c}{\includegraphics[width=\textwidth]{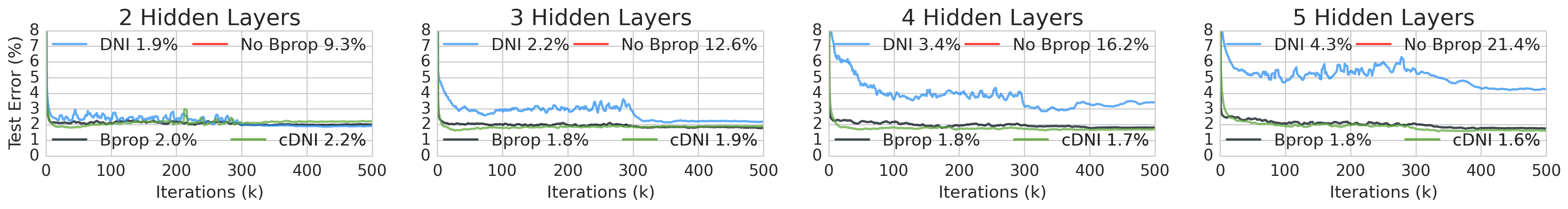}}\\
\multicolumn{2}{c}{\small CIFAR-10 FCN}\\
\multicolumn{2}{c}{\includegraphics[width=\textwidth]{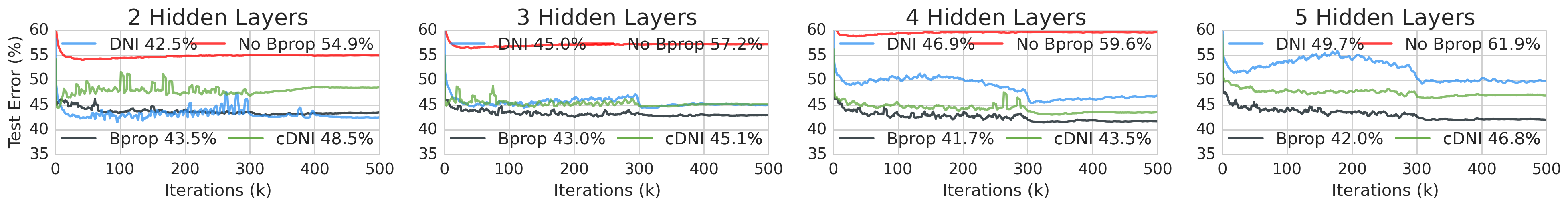}}\\
~~~~~~~~~~~~~~~~~~~~~~~~~~~~~~~~~~~\small MNIST CNN&
~~~~~~~~~~~~~~~~~~~~~~~~~~~~~~\small CIFAR-10 CNN\\
\multicolumn{2}{c}{\includegraphics[width=\textwidth]{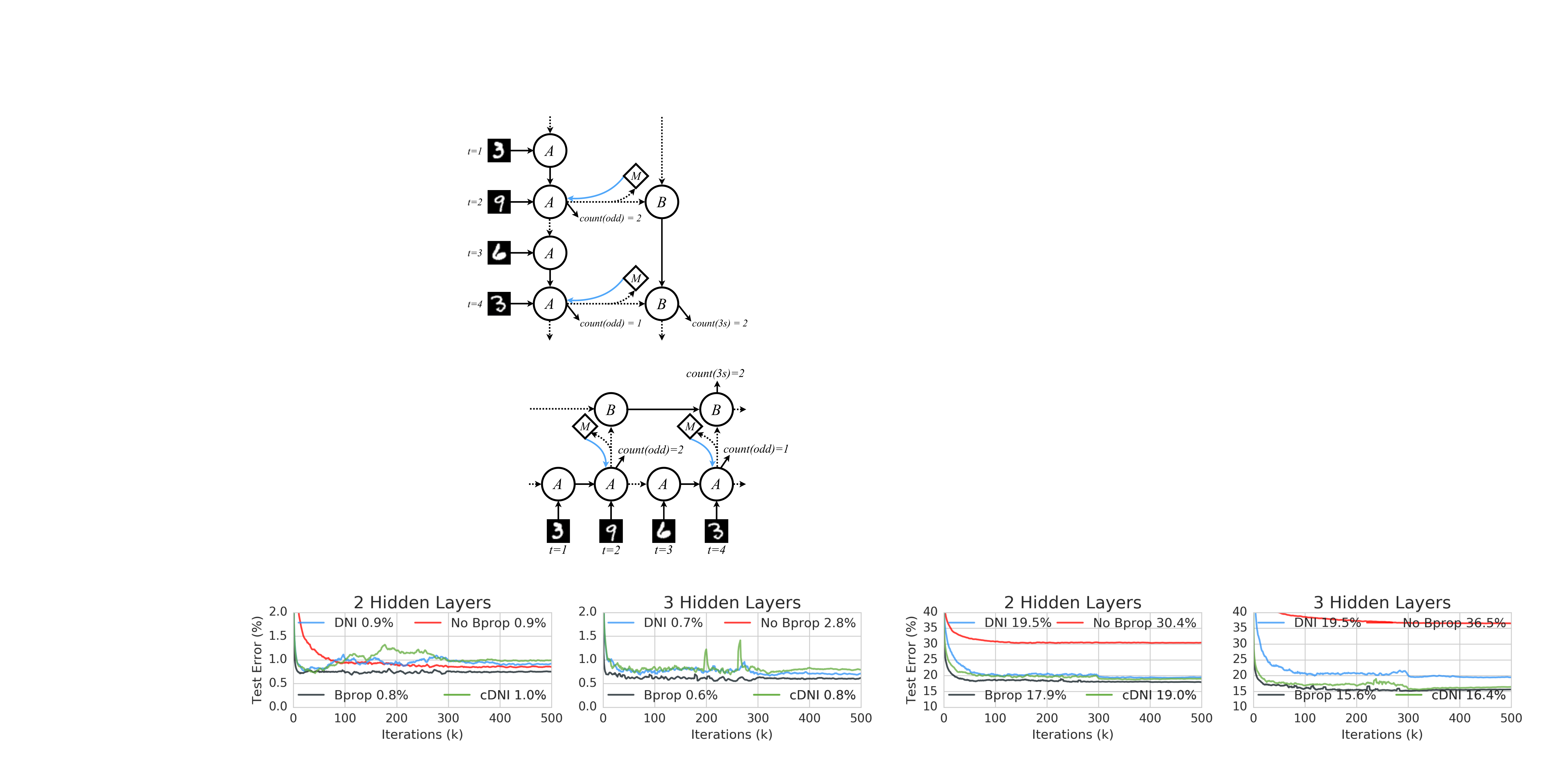}}\\
\end{tabular}
\caption{\small Corresponding test error curves during training for the results in \tblref{table:ff}. (a) MNIST digit classification with FCNs, (b) CIFAR-10 image classification with FCNs. DNI can be easily used with CNNs as shown in (c) for CNNs on MNIST and (d) for CNNs on CIFAR-10.}
\label{fig:fcnall}
\end{figure*}

\begin{figure*}[t]
\centering
\begin{tabular}{cccc}
\hspace{-4em}
\includegraphics[width=\textwidth]{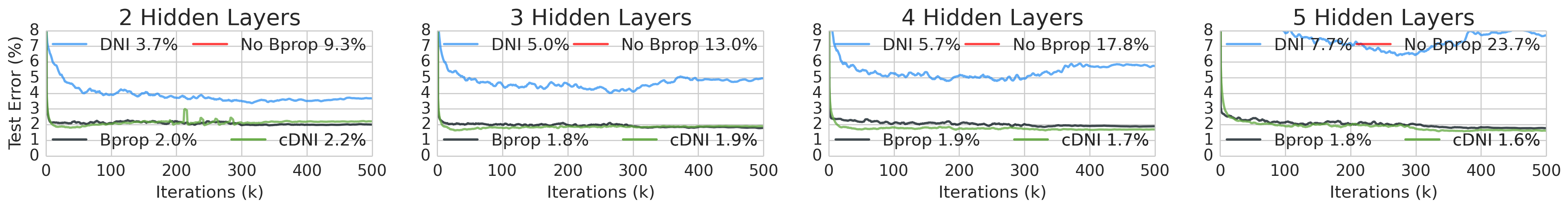}\\
\end{tabular}
\caption{\small Linear DNI models for FCNs on MNIST.}
\label{fig:mnistlinear}
\end{figure*}

\subsection{Underfitting of Synthetic Gradient Models}\label{sec:appunderfitting}
\begin{figure*}[t]
\centering
\begin{tabular}{cccc}
\includegraphics[width=0.6\textwidth]{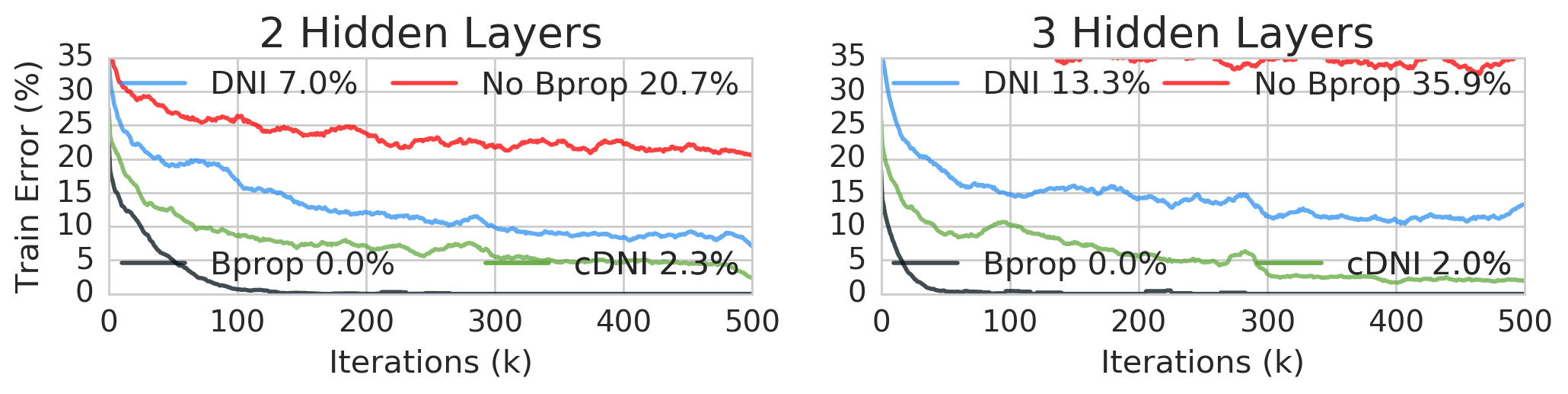}\\
(a)
\end{tabular}
\caption{\small (a) Training error for CIFAR-10 CNNs.}
\label{fig:cifartrain}
\end{figure*}

If one takes a closer look at learning curves for DNI model (see \figref{fig:cifartrain} for training error plot on CIFAR-10 with CNN model) it is easy to notice that the large test error (and its degradation with depth) is actually an effect of underfitting and not lack of ability to generalise or lack of convergence of learning process. One of the possible explanations is the fact that due to lack of label signal in the DNI module, the network is over-regularised as in each iteration DNI tries to model an expected gradient over the label distribution. This is obviously a harder problem than modelling actual gradient, and due to underfitting to this subproblem, the whole network also underfits to the problem at hand. Once label information is introduced in the cDNI model, the network fits the training data much better, however using synthetic gradients still acts like a regulariser, which also translates to a reduced test error. This might also suggest, that the proposed method of conditioning on labels can be further modified to reduce the underfitting effect.

\begin{figure*}[t]
\centering
\begin{tabular}{cccc}
\includegraphics[width=0.8\textwidth]{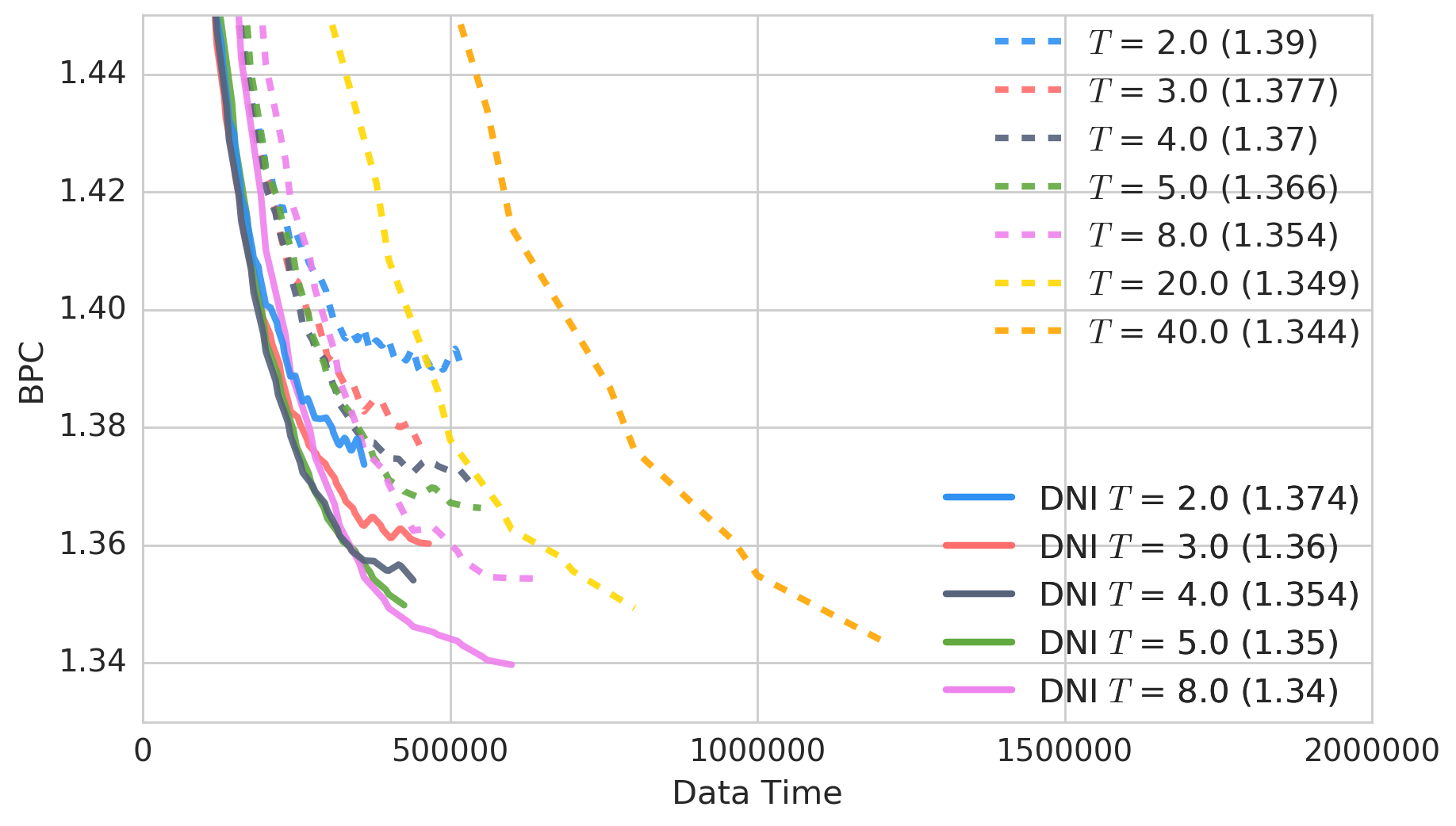}\\
\includegraphics[width=0.8\textwidth]{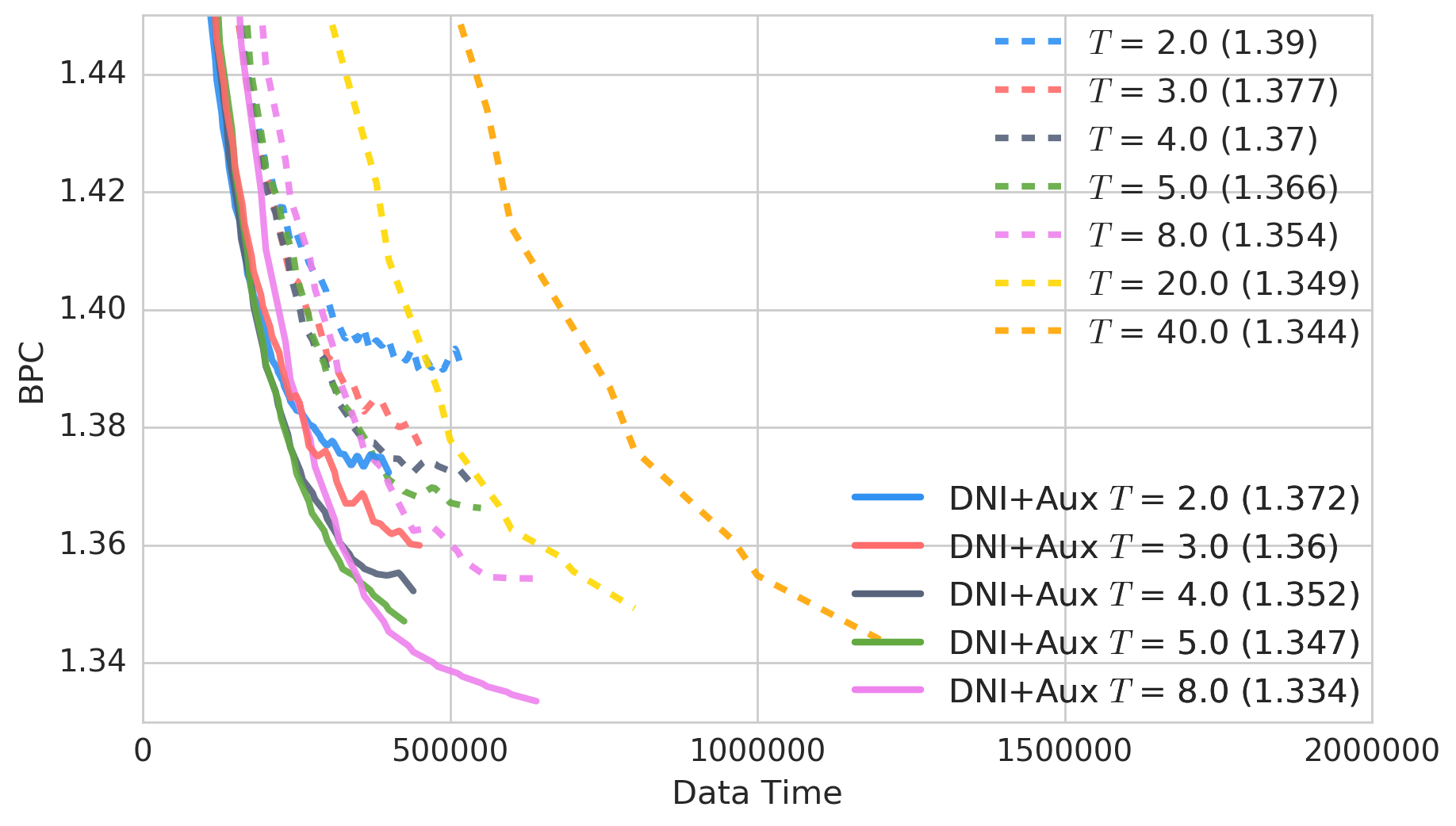}\\
\end{tabular}
\caption{\small Test error in bits per character (BPC) for Penn Treebank character modelling. We train the RNNs with different BPTT unroll lengths with DNI (solid lines) and without DNI (dashed lines). Early stopping is performed based on the validation set. Top shows results with DNI, and bottom shows results with DNI and future synthetic gradient prediction (DNI+Aux). Bracketed numbers give final test set BPC.}
\label{fig:ptb}
\end{figure*}

\begin{figure*}[t]
\centering
\begin{tabular}{cccc}
\includegraphics[width=1.0\textwidth]{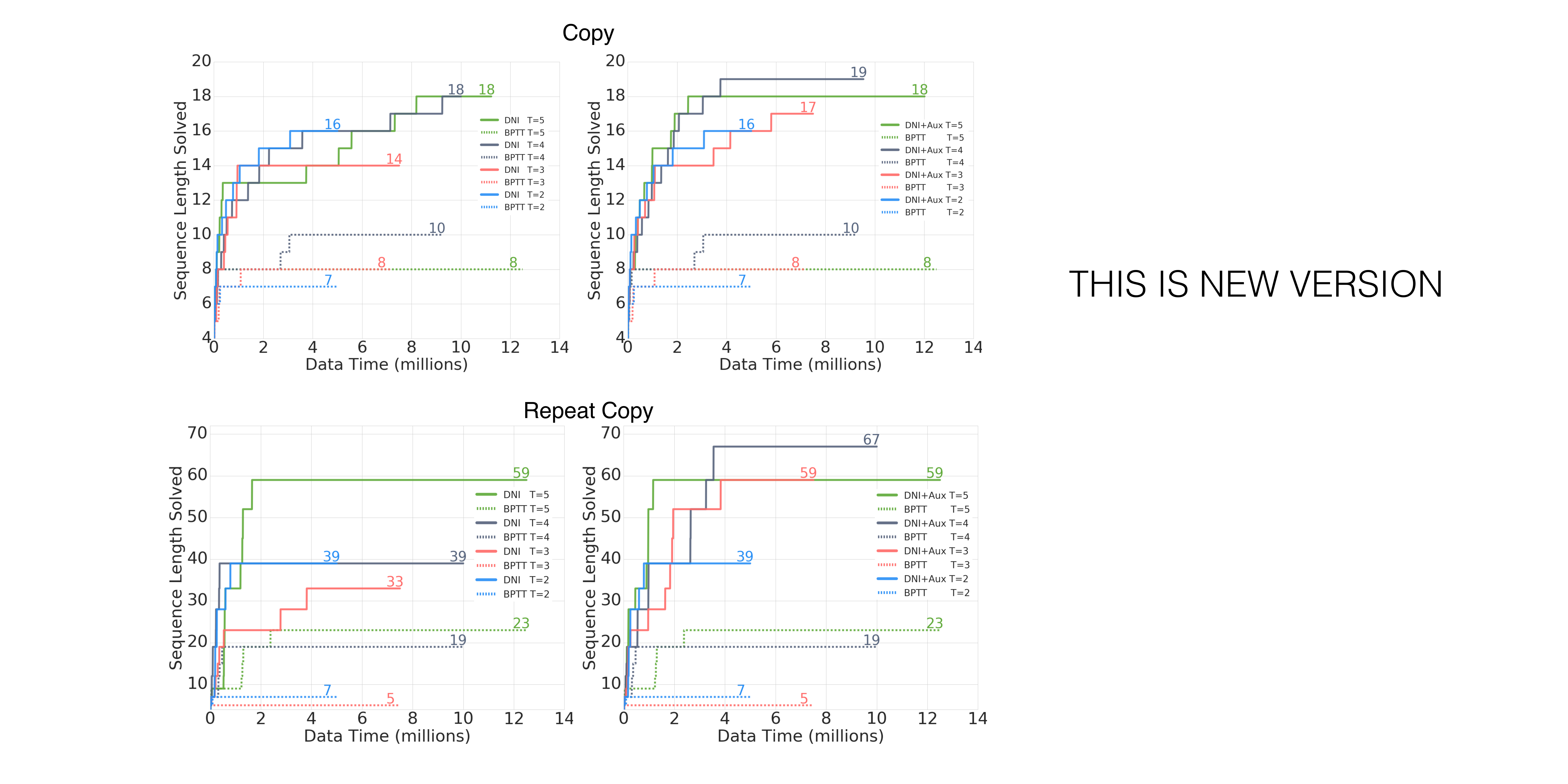}
\end{tabular}
\caption{\small The task progression for Copy (top row) and Repeat Copy (bottom row) without future synthetic gradient prediction (left) and with future synthetic gradient prediction (right). For all experiments the tasks' time dependency is advanced after the RNN reaches 0.15 bits error. We run all models for 2.5M optimisation steps. The x-axis shows the number of samples consumed by the model, and the y-axis the time dependency level solved by the model -- step changes in the time dependency indicate that a particular time dependency is deemed solved. DNI+Aux refers to DNI with the additional future synthetic gradient prediction auxiliary task.}
\label{fig:copyloop}
\end{figure*}

\section{Implementation Details}

\subsection{Feed-Forward Implementation Details}\label{sec:appffdetails}
In this section we give the implementation details of the experimental setup used in the experiments from \sref{sec:expff}. 

\paragraph{Conditional DNI (cDNI)}
In order to provide DNI module with the label information in FCN, we simply concatenate the one-hot representation of a sample's label to the input of the synthetic gradient model. Consequently for both MNIST and CIFAR-10 experiments, each cDNI module takes ten additional, binary inputs. For convolutional networks we add label information in the form of one-hot encoded channel masks, thus we simply concatenate ten additional channels to the activations, nine out of which are filled with zeros, and one (corresponding to sample's label) is filled with ones.

\paragraph{Common Details}
All experiments are run for 500k iterations and optimised with Adam~\citep{Kingma14} with batch size of 256. The learning rate was initialised at $3\times 10^{-5}$ and decreased by a factor of 10 at 300k and 400k steps. Note the number of iterations, learning rate, and learning rate schedule was not optimised. We perform a hyperparameter search over the number of hidden layers in the synthetic gradient model (from 0 to 2, where 0 means we use a linear model such that $\hat{\delta} = M(h) = \phi_w h + \phi_b$) and select the best number of layers for each experiment type (given below) based on the final test performance. We used cross entropy loss for classification and L$_2$ loss for synthetic gradient regression which was weighted by a factor of 1 with respect to the classification loss. All input data was scaled to $[0, 1]$ interval. The final regression layer of all synthetic gradient models are initialised with zero weights and biases, so initially, zero synthetic gradient is produced.

\paragraph{MNIST FCN} 
Every hidden layer consists of fully-connected layers with 256 units, followed by batch-normalisation and ReLU non-linearity. The synthetic gradient models consists of two (DNI) or zero (cDNI) hidden layers and with 1024 units (linear, batch-normalisation, ReLU) followed by a final linear layer with 256 units.

\paragraph{MNIST CNN}
The hidden layers are all convolutional layers with 64 $5\times 5$ filters with resolution preserving padding, followed by batch-normalisation, ReLU and $3 \times 3$ spatial max-pooling in the first layer and average-pooling in the remaining ones. The synthetic gradient model has two hidden layers with 64 $5\times 5$ filters with resolution preserving padding, batch-normalisation and ReLU, followed by a final 64 $5 \times 5$ filter convolutional layer with resolution preserving padding.

\paragraph{CIFAR-10 FCN}
Every hidden layer consists of fully-connected layers with 1000 units, followed by batch-normalisation and ReLU non-linearity. The synthetic gradient models consisted of one hidden layer with 4000 units (linear, batch-normalisation, ReLU) followed by a final linear layer with 1000 units.

\paragraph{CIFAR-10 CNN}
The hidden layers are all convolutional layers with 128 $5\times 5$ filters with resolution preserving padding, followed by batch-normalisation, ReLU and $3 \times 3$ spatial max-pooling in the first layer and avg-pooling in the remaining ones. The synthetic gradient model has two hidden layers with 128 $5\times 5$ filters with resolution preserving padding, batch-normalisation and ReLU, followed by a final 128 $5 \times 5$ filter convolutional layer with resolution preserving padding.

\paragraph{Complete Unlock.}
In the completely unlocked model, we use the identical architecture used for the synthetic gradient model, but for simplicity both synthetic gradient and synthetic input models use a single hidden layer (for both DNI and cDNI), and train it to produce synthetic inputs $\hat{h}_i$ such that $\hat{h}_i \simeq h_i$. The overall training setup is depicted in \figref{fig:drasticnet}. During testing all layers are connected to each other for a forward pass, \ie~the synthetic inputs are not used.

\subsection{RNN Implementation Details}\label{sec:apprnndetails}
\paragraph{Common Details}
All RNN experiments are performed with an LSTM recurrent core, where the output is used for a final linear layer to model the task. In the case of DNI and DNI+Aux, the output of the LSTM is also used as input to a single hidden layer synthetic gradient model with the same number of units as the LSTM, with a final linear projection to two times the number of units of the LSTM (to produce the synthetic gradient of the output and the cell state). The synthetic gradient is scaled by a factor of 0.1 when consumed by the model (we found that this reliably leads to stable training). We perform a hyperparameter search of whether or not to backpropagate synthetic gradient model error into the LSTM core (the model was not particularly sensitive to this, but occasionally backpropagating synthetic gradient model error resulted in more unstable training). The cost on the synthetic gradient regression loss and future synthetic gradient regression loss is simply weighted by a factor of 1.

\paragraph{Copy and Repeat Copy Task}
In these tasks we use 256 LSTM units and the model was optimised with Adam with a learning rate of $7\times 10^{-5}$ and a batch size of 256. The tasks were progressed to a longer episode length after a model gets below 0.15 bits error. The Copy task was progressed by incrementing $N$, the length of the sequence to copy, by one. The Repeat Copy task was progressed by alternating incrementing $N$ by one and $R$, the number of times to repeat, by one.

\paragraph{Penn Treebank}
The architecture used for Penn Treebank experiments consists of an LSTM with 1024 units trained on a character-level language modelling task.
Learning is performed with the use of Adam with learning rate of $7\times 10^-5$ (which we select to maximise the score of the baseline model through testing also $1\times 10^{-4}$ and $1\times 10^{-6}$) without any learning rate decay
or additional regularisation. Each 5k iterations we record validation error (in terms of average bytes per character) and store the network which achieved the smallest one. Once validation error starts to increase we stop training and report test error using previously saved network. In other words, test error is reported for the model yielding minimum validation error measured with 5k iterations
resolution. A single iteration consists of performing full BPTT over $T$ steps with a batch of 256 samples. 

\subsection{Multi-Network Implementation Details}\label{sec:appmultidetails}

The two RNNs in this experiment, Network A and Network B, are both LSTMs with 256 units which use batch-normalisation as described in \citep{Cooijmans16}. Network A takes a $28\times 28$ MNIST digit as input and has a two layer FCN (each layer having 256 units and consisting of linear, batch-normalisation, and ReLU), the output of which is passed as input to its LSTM. The output of Network A's LSTM is used by a linear classification layer to classify the number of odd numbers, as well as input to another linear layer with batch-normalisation which produces the message to send to Network B. Network B takes the message from Network A as input to its LSTM, and uses the output of its LSTM for a linear classifier to classify the number of 3's seen in Network A's datastream. The synthetic gradient model has a single hidden layer of size 256 followed by a linear layer which produces the 256-dimensional synthetic gradient as feedback to Network A's message.

All networks are trained with Adam with a learning rate of $1\times 10^{-5}$. We performed a hyperparameter search over the factor by which the synthetic gradient should by multiplied by before being backpropagated through Network A, which we selected as 10 by choosing the system with the lowest training error.

 \bibliography{shortstrings,vgg_local,vgg_other,current}
 \bibliographystyle{icml2017}


\end{appendices}

\end{document}